\documentclass[acmtog]{acmart}

\AtBeginDocument{%
  }

\setcopyright{acmlicensed}
\copyrightyear{2024}
\acmYear{2024}
\acmDOI{XXXXXXX.XXXXXXX}
\acmJournal{TOG}
\acmVolume{1}
\acmNumber{1}
\acmArticle{1}
\acmMonth{10}

\usepackage{color}

\newcommand{\modi}[1]{{\color{black} #1}}
\newcommand{\re}[1]{{\color{black} #1}}

\usepackage{wrapfig}
\usepackage{graphicx}
\usepackage{overpic}
\usepackage{array}
\usepackage{booktabs}
\usepackage{amsmath}
\usepackage{amsthm}

\usepackage{amssymb}
\usepackage{makebox}
\usepackage{rotating}
\usepackage{makecell}
\usepackage{multirow}
\newtheorem{Def}{Definition}

\begin{document}


\title{Quasi-Medial Distance Field (Q-MDF): A Robust Method for Approximating and Discretizing Neural Medial Axes }


\author{Jiayi Kong}
\authornote{J. Kong and C. Zong contributed equally to this research.}
\email{jiayi006@e.ntu.edu.sg}
 \orcid{0009-0004-6922-2354}
  \affiliation{%
  \institution{S-Lab, Nanyang Technological University}
  \country{Singapore}
 }
 
 \author{Chen Zong}
 \email{zongchen@nuaa.edu.cn}
 \orcid{0000-0003-4954-0780}
 \authornotemark[1]
  \affiliation{%
  \institution{S-Lab, Nanyang Technological University}
  \country{Singapore}
 }
 \affiliation{
 \institution{School of Mathematics, Nanjing University of Aeronautics and Astronautics}
 \country{China}
 }
 
\author{Jun Luo}
\email{junluo@ntu.edu.sg}
\orcid{0000-0002-7036-5158}
 \affiliation{%
  \institution{S-Lab, Nanyang Technological University}
  \country{Singapore}
 }

\author{Shiqing Xin}
\email{xinshiqing@sdu.edu.cn}
\orcid{0000-0001-8452-8723}
\affiliation{
\institution{School of Computer Science and Technology, Shandong University}
\city{Qingdao}
\country{China}
}

\author{Fei Hou}
\email{houfei@ios.ac.cn}
\orcid{0000-0001-8226-6635}
\affiliation{
\institution{Key Laboratory of System Software (CAS), Institute of Software, Chinese Academy of Sciences, and University of Chinese Academy of Sciences}
\city{Beijing}
\country{China}
}

\author{Hanqing Jiang}
\email{jianghanqing@sensetime.com}
\orcid{0000-0001-9582-5539}
\author{Chen Qian}
\email{qianchen@sensetime.com}
\orcid{0000-0002-8761-5563}
\affiliation{\institution{SenseTime Research}
\country{China}}

\author{Ying He}
\authornote{Corresponding author: Y. He.}
\email{yhe@ntu.edu.sg}
\orcid{0000-0002-6749-4485}
 \affiliation{%
  \institution{S-Lab, Nanyang Technological University}
  \country{Singapore}
 }




\begin{abstract}
The medial axis, a lower-dimensional descriptor that captures the extrinsic structure of a shape, plays an important role in digital geometry processing. Despite its importance, computing the medial axis transform robustly from diverse inputs, especially point clouds with defects, remains a challenging problem.
In this paper, we propose a new implicit method that deviates from traditional explicit medial axis computation. 
Our key technical insight is that the difference between the signed distance field (SDF) and the medial field (MF) of a solid shape relates to the unsigned distance field (UDF) of the shape's medial axis. This observation allows us to  formulate medial axis extraction as an implicit reconstruction problem. By employing a modified double covering strategy, we recover the medial axis as the zero level-set of the UDF.
Extensive experiments demonstrate that our method achieves higher accuracy and robustness in learning compact medial axis transforms from challenging meshes and point clouds, outperforming existing approaches.
\end{abstract}




\begin{teaserfigure}
\centering
\includegraphics[width=\textwidth]{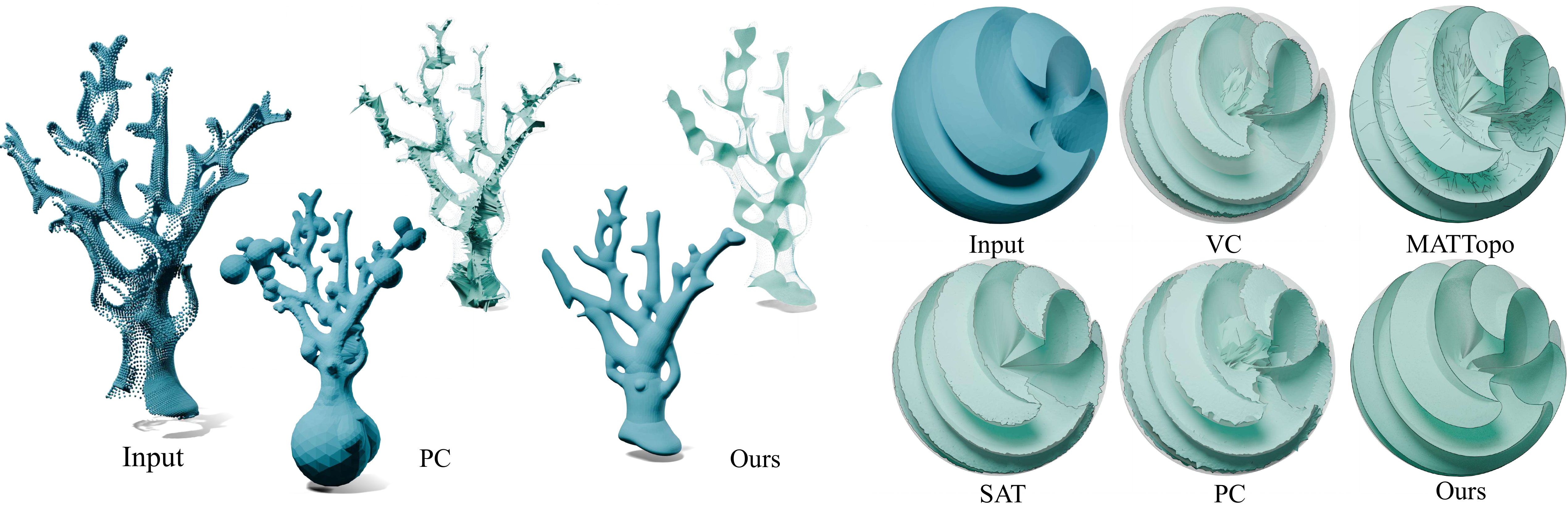}
\caption{We introduce a differentiable field, Q-MDF, for robust medial axis approximation, from which a medial mesh can be directly extracted. Compared with existing methods such as PC~\cite{amenta2001power}, VC~\cite{yan2018voxel}, MATTopo~\cite{wang2024mattopo}, and SAT~\cite{miklos2010discrete}, Q-MDF demonstrates enhanced robustness
in computing compact medial axis from diverse inputs. \textbf{Left:} The coral model contains missing regions and non-uniform point distribution, which challenge PC and lead to an off-centered media axis. The surface reconstructed from such an axis tends to suffer from geometric and topological artifacts. 
In contrast, Q-MDF produces a stable medial axis and a clean reconstructed surface. \textbf{Right:} \modi{A sphere-like mesh with sharp features and the corresponding medial axes produced by Q-MDF and the baselines. Surface boundaries are shown in black.} Q-MDF yields a more compact medial axis while faithfully preserving sharp features.
}
\label{fig:teaser}
\end{teaserfigure}
\maketitle

\section{Introduction}
\label{sec:introduction}

The medial axis (MA)~\cite{Blum1967} is a classical geometric structure that encapsulates both the local symmetries and global topological characteristics of a shape. When augmented with radius information, known as the medial axis transform (MAT), it enables the reconstruction of the original shape and serves as an essential lower-dimensional shape descriptor. The medial axis has found extensive applications in digital geometry processing, including shape synthesis and design~\cite{tang2019skeleton,petrov2024gem3d,noma2024surface}, segmentation~\cite{zhou2015generalized,lin2020seg}, pose analysis~\cite{yang2021learning,dou2023tore}, and animation~\cite{yang2018dmat,lan2020medial,lan2021medial}, among others.

As a shape representation, the medial axis lies at the geometric center of an object, effectively capturing its protrusions and structural components. Topologically, it is homotopy equivalent to the shape under standard smoothness assumptions. Despite these appealing properties, the medial axis is notoriously unstable; even slight boundary perturbations can create spurious branches, as illustrated in Figure~\ref{fig:ma_2D}~(c), thereby increasing its complexity and reducing its usability in downstream applications. Moreover, input data are not always watertight or densely sampled. In practice, point clouds often contain defects such as noise, non-uniform sampling, and missing regions, making it difficult to  infer a plausible shape. Although the medial axis is rigorously defined, computing a compact and stable version from such imperfect data remains a long-standing challenge.



Most \modi{traditional} medial axis computation techniques rely on discretization \modi{of either the shape or the medial axis itself, such as voxelization or direct extraction of medial points~\cite{saha2016survey,tagliasacchi20163d}.} Voxel-based methods identify a subset of voxels that approximately satisfy medial axis criteria. They are simple and efficient, but they are inherently limited by resolution, which constrains accuracy and scalability. Medial point-based approaches represent the medial axis as a set of points with connectivity, often derived from subsets of the Voronoi diagram constructed from sampled surface points. These methods aim to balance compactness with reconstruction accuracy by simplifying the Voronoi structure or grouping symmetric point pairs to form a medial mesh. However, discretization-based methods typically require high-quality inputs (e.g., watertight meshes or dense, noise-free point clouds) and often produce suboptimal results when these assumptions fail.

Recent learning-based approaches alleviate some input constraints. Among them, \citet{rebain2021deep} made a seminal contribution by introducing the \emph{medial field} (MF), a continuous and differentiable representation that models the radius information of the medial axis. They jointly trained the medial field with the signed distance field (SDF) within a unified neural framework. This elegant formulation not only provides a differentiable characterization medial geometry but also enables practical applications such as accelerating ray-surface intersection computations. However, because the medial field primarily encodes the radius information of  projected surface points, it does not explicitly represent the medial axis itself, leaving open the problem of recovering a compact and accurate medial structure.

Inspired by the insights in~\cite{rebain2021deep}, we propose a new neural implicit representation of the medial axis, from which a compact and robust medial mesh can be derived. Our key observation is that, for a given solid shape, the difference between its SDF and MF forms a differentiable function analogous to an unsigned distance field (UDF) of the medial axis. Consequently, the computation of the medial axis can be formulated as a UDF learning problem. 

Extracting a non-manifold surface from such a UDF, however, is challenging because the UDF lacks sign information to distinguish between interior and exterior regions. Standard isosurface extraction methods, such as Marching Cubes~\cite{lorensen1998marching} or Dual Contouring~\cite{Ju2002}, therefore fail to recover consistent surface topology or orientation, particularly in non-manifold regions where multiple medial sheets intersect or merge. 

Following the idea of DCUDF~\cite{hou2023robust}, 
we adopt a modified double covering strategy to capture the structure of the medial axis by creating a two-layered enclosure. By collapsing the volume between the two layers to zero, we obtain a compact medial mesh that faithfully represents the medial surface. Furthermore, owing to the feature-sensitive nature of the medial axis, we enhance the learned neural implicit representation near sharp regions, resulting in a more precise and refined medial structure.

Experimental results show that our formulation broadens the applicability of the original medial field~\cite{rebain2021deep}, enabling the computation of simplified medial axes even in challenging scenarios and  providing a new perspective on neural shape representation.

Our main contributions are summarized as follows:
\begin{itemize}
    \item We develop a robust approach for approximating the medial axis transform that takes imperfect or complex data as input and reliably produces a compact and plausible medial axis representation. 
    \item We establish a theoretical link between the signed distance field and the medial field, which leads to a neural implicit representation of the medial axis.
    \item We introduce an effective double covering-based algorithm to accurately extract a medial mesh from the learned continuous representation. 
    \item We improve the learned field near sharp regions by using the ``pointing-to-convex-feature'' property of the medial axis, which enhances feature preservation and precision.
\end{itemize}

\section{Related Work}
\label{sec:relatedworks}

The medial axis problem has been extensively studied over the past few decades, resulting in a large body of literature. Due to space limitations, this section reviews only the most recent and relevant methods. For a more comprehensive literature review, we refer readers to \cite{tagliasacchi20163d}.

\subsection{Traditional Approaches}
\label{subsec:traditionalmethods}

\paragraph{Analytical Approaches}
Analytical approaches aim to compute the medial axis according to its mathematical definition. A classic idea starts with the analysis of localized shape boundaries, tracing seam curves and identifying junction points to construct the medial axis. Representative methods include those designed for polyhedra~\cite{Culver2004exact,milenkovic1993robust,sherbrooke1996algorithm,Culver1999accurate} and for simple free-form shapes~\cite{Ramanathan2010interior}. Due to their strict requirements for input simplicity, these methods have limited practical use.

\paragraph{Voxel-based Approaches}
Most voxel-based approaches discretize the target shape into a voxel grid and search for a subset that satisfies medial axis criteria (e.g., thinness, centrality, and homotopy equivalence) based on certain distance measures~\cite{saha2016survey}.
Examples include methods using Manhattan distance~\cite{palagyi1999parallel}, Chamfer distance~\cite{pudney1998distance}, and Euclidean distance~\cite{hesselink2008euclidean,rumpf2002continuous}. 
\citet{yan2018voxel} further provided a rigorous sampling condition using voxel representation and constructed a medial mesh by taking the dual of the interior Delaunay triangulation, which is theoretically guaranteed to be topologically equivalent to the original shape.

\paragraph{Approximation with Paired Points}
This family of methods begins with a set of oriented surface points and approximates medial points as centers of locally maximum inscribed spheres~\cite{ma20123d}. The connectivity among medial points is then established through localized plane fitting~\cite{jalba2012surface}. These approaches provide a coarse medial description of the shape, often sacrificing geometric accuracy and topological correctness.

\paragraph{Voronoi-guided Approaches}
The Voronoi diagram has played a major role in medial axis computation due to its spatial partitioning property and computational efficiency. 
It was proven that in $\mathbb{R}^2$, if the boundary curve satisfies an $\epsilon$-sampling condition, the subset of the Voronoi diagram fully contained within the shape approximates the medial axis~\cite{brandt1994convergence}.
However, this result could not be directly extended to 3D due to the presence of ``slivers,'' which  produce Voronoi vertices close to the surface and create redundant branches~\cite{amenta2001power}.
As a result, many studies focus on removing these redundant branches.
\citet{attali1996modeling} and \citet{dey2004approximating} applied an angle criterion, although topology was not always preserved. \citet{giesen2006medial} improved topological correctness by using unstable manifolds of the Voronoi diagram.
\citet{chazal2005lambda} and \citet{chazal2008smooth} employed a radius criterion that preserves topology but overlooks geometric fidelity. \citet{miklos2010discrete} scaled medial balls and filtered out vertices whose corresponding balls are absorbed by others.
\citet{li2015q} and \citet{pan2019q} proposed importance measures to iteratively collapse less important edges and generate simplified medial points using a quadratic error metric.

Another class of methods obtains medial points first and then leverages weighted Voronoi properties to establish connectivity.
\citet{amenta2001power} selected a subset of Voronoi vertices as ``poles'' and used weighted Voronoi diagrams to infer connectivity, forming a structure known as the power shape, which converges to the medial axis as sampling density increases. 
To obtain compact representations, \citet{dou2022coverage} and \citet{wang2024coverage} incrementally selected skeletal points from candidate sets until a target number was reached, inferring connectivity from the primitive medial surface.
\citet{wang2022computing} proposed a feature-preserving method for CAD-like models by optimizing  inner medial balls via a restricted power diagram. More recently, 
\citet{wang2024mattopo} further improved topological correctness while maintaining geometric fidelity, although extensive tetrahedral slicing operations increase computational overhead.

\subsection{Learning-based Approaches}
\label{subsec:learningmethods}
Most learning-based methods adopt discrete formulations to identify medial points and infer their connectivity. \citet{Yang2020p2mat} computed medial points with radius information from sparse point clouds via point-wise displacement vectors and applied tetrahedral simplification to construct  medial meshes.
\citet{lin2021point2skeleton} and \citet{ge2023point2mm} learned medial points with associated radii using graph auto-encoder networks to establish connectivity. However, these methods often suffer from limited reconstruction accuracy and incorrect topology. 
\citet{Clemot2023neural} approached the problem from a field-based perspective, incorporating total variation regularization to improve the quality of SDFs. They traced medial points from surface samples and formulated a mixed-integer linear programming problem to compute a simplified medial axis, achieving robustness against noisy and incomplete point clouds.

\citet{rebain2021deep} explored a differentiable field representation, termed the \emph{medial field}, and jointly trained it with the SDF to improve the accuracy of both fields. While their method shows clear benefits for accelerating ray-surface intersection calculations, they did not investigate its potential for direct medial mesh computation. This paper can be viewed as an immediate follow-up to~\cite{rebain2021deep}, explicitly addressing the problem of medial axis extraction and demonstrating its effectiveness in challenging scenarios where traditional methods often fail.

\section{Preliminaries}
\label{sec:preliminaries}
Let $\Omega\subset\mathbb{R}^3$ be a solid 3D shape with boundary surface $\partial\Omega$. We denote the set of interior points by $\Omega^-$ and the set of exterior points by $\Omega^+$. This section briefly introduces the key concepts related to the medial axis, the signed distance field, and the unsigned distance field defined on $\Omega$, and extends these notations to the implicit representation of the medial axis.

\subsection{Terminologies}\label{subsec:terminologies}
A point $p\in\mathbb{R}^3$ is called a \textbf{medial point} if it has more than one closest point on the boundary $\partial\Omega$. The distance between $p$ and any of its closest boundary points is referred to as the \textbf{radius}. The \textbf{medial axis}  $\mathcal{M}$ is the set of all medial points and can be viewed as the locus of centers of maximally inscribed spheres within the shape.

The medial axis can be divided into two parts: the portion lying inside the shape, referred to as the \textbf{inner medial axis}, and the portion outside, referred to as the \textbf{outer medial axis}. When the medial axis is associated with the radius of its corresponding maximal inscribed sphere, the structure is known as the \textbf{medial axis transform}.

In this paper, we focus on computing the inner medial axis transform of $\Omega$, which we denote as $\mathcal{M}_{\Omega}$ and simply call the ``medial axis''. From the perspective of the continuous distance field, medial points correspond to the local minima of the signed distance function of the shape. 
Similar to \cite{rebain2021deep}, for any point $p\in\Omega^-$, its \textbf{projected medial point}, denoted $\mathcal{M}_{\text{proj}}(p)$, is defined
as the intersection between the medial axis and the ray originating from $p$ and directed along the inward normal of its closest surface point on $\partial\Omega$. 

\begin{figure}[htb]
    \centering
    \includegraphics[width=0.31\linewidth]{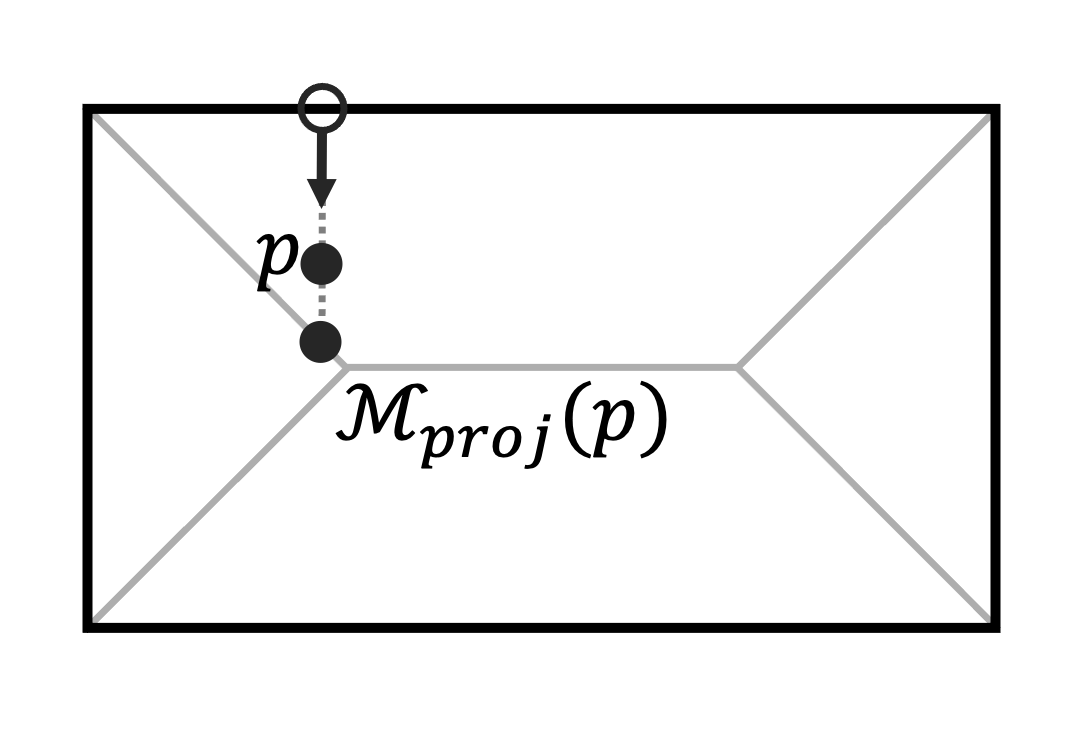}
    \includegraphics[width=0.31\linewidth]{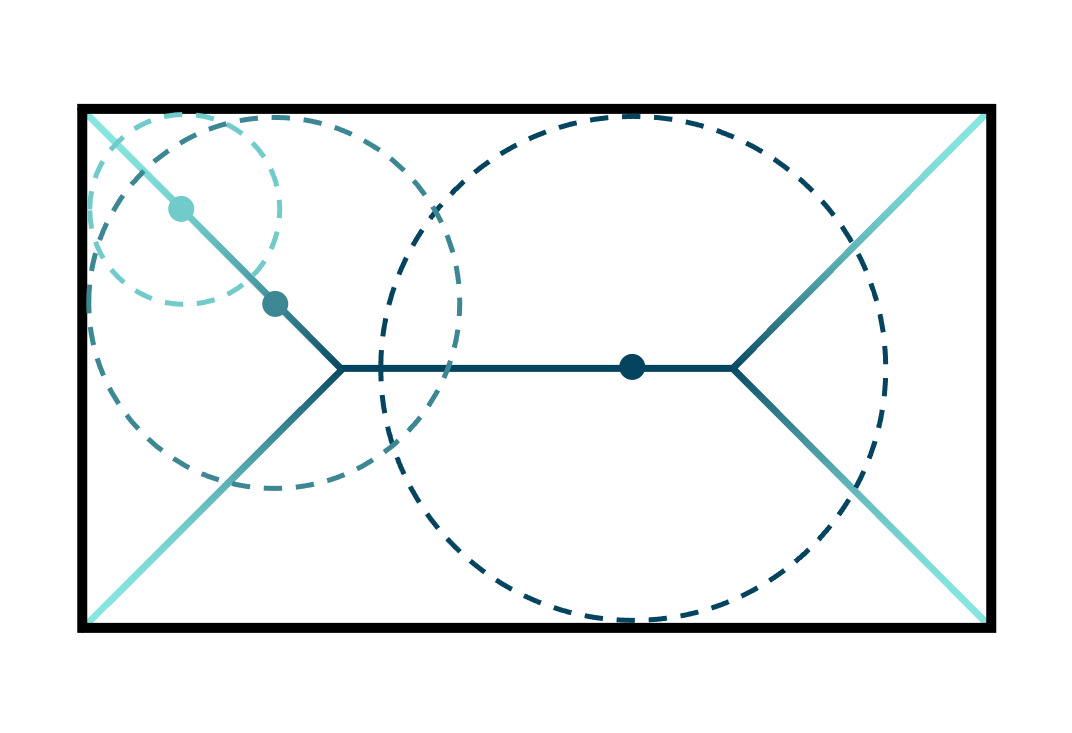}
    \includegraphics[width=0.31\linewidth]{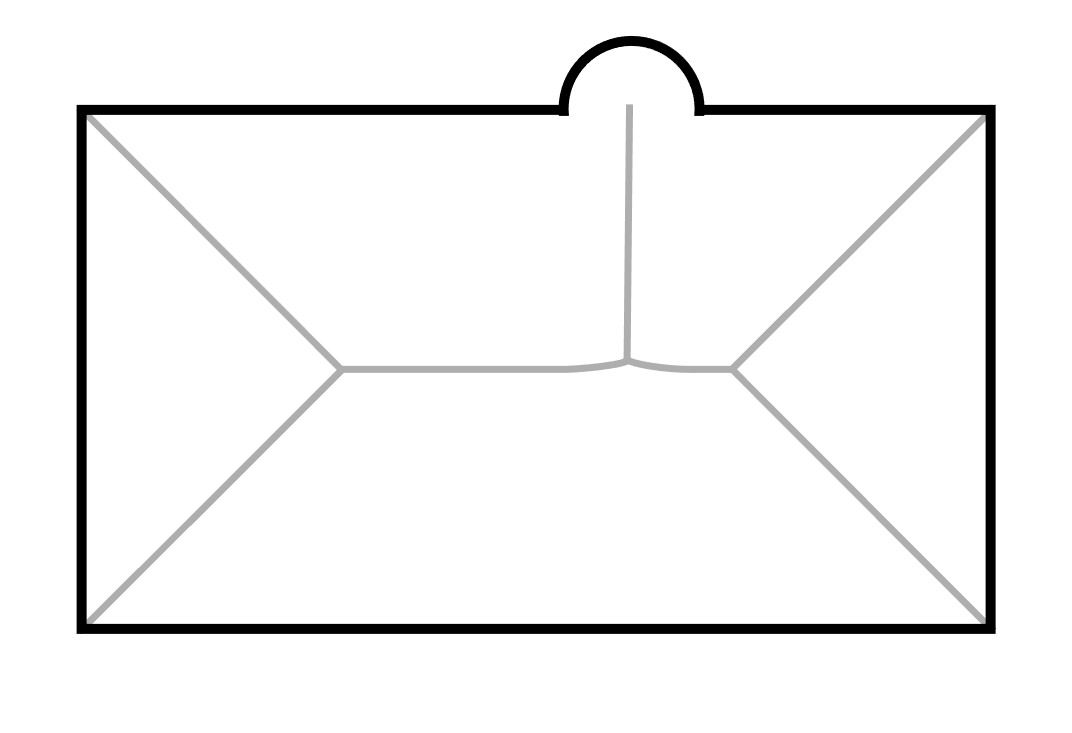}
    \makebox[0.31\linewidth][c]{(a) MA}
    \makebox[0.31\linewidth][c]{(b) MAT}
    \makebox[0.31\linewidth][c]{(c) Unstability}
    \caption{2D illustration of the medial axis and medial axis transform. The medial axis, a lower-dimensional and inherently unstable representation of the original shape, is highly sensitive to boundary perturbations that can significantly alter its topology.}
    \label{fig:ma_2D}
\end{figure}

The medial axis $\mathcal{M}$ can be regarded as a lower-dimensional representation of $\Omega$, since its geometric structure forms a non-manifold curved surface rather than a volumetric solid. It allows for the exact reconstruction of the original shape by sweeping spheres of varying radii along its structure. However, the medial axis is inherently unstable: even small perturbations on the boundary can introduce additional branches and dramatically increase its topological complexity, as illustrated in  Figure~\ref{fig:ma_2D}~(c).

\subsection{Distance Fields for Implicit Surface Representation}
\label{subsec:preliminary-implicitshaperepresentation}

A distance field provides an efficient and powerful way to represent a shape implicitly.
In this section, we briefly review two commonly used variants:  \textbf{signed distance field} and \textbf{unsigned distance field}.

\begin{figure}[htb]
    \centering
    \includegraphics[width=\linewidth]{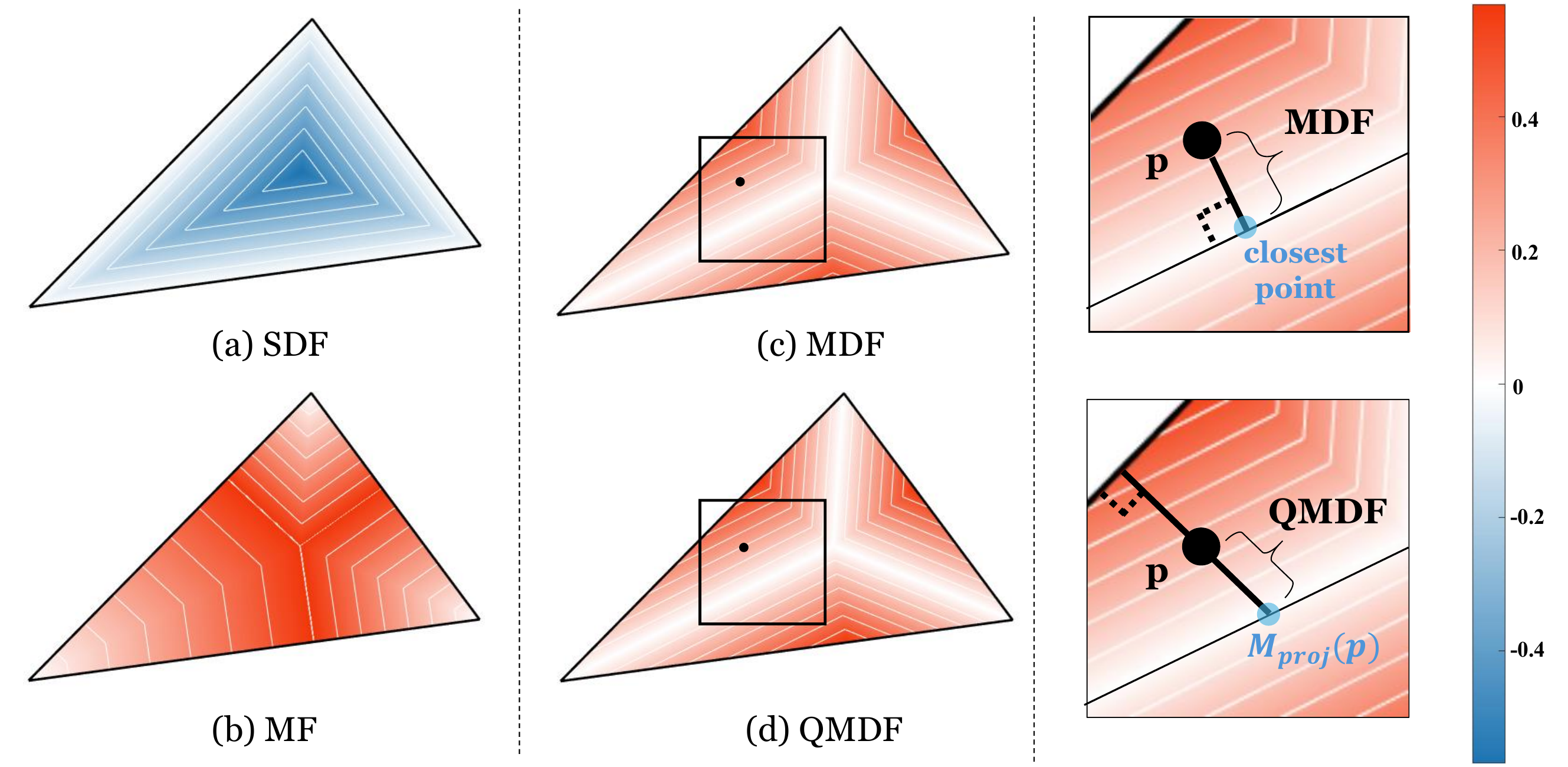}
    \caption{Distance fields and medial fields. Given a 2D triangle, (a) visualizes the signed distance field,  and (b) the medial field. In (c), the medial distance field (MDF), which is defined as the UDF of the medial axis. (d) illustrates the quasi-medial distance field (Q-MDF), computed by subtracting the SDF from the MF, which effectively approximates the medial distance field.}
    \label{fig:field2D}
\end{figure}

\paragraph{Signed Distance Field} 
 The minimum distance between an arbitrary point $p\in \mathbb{R}^3$ and the boundary $\partial\Omega$ is defined as 
$$d(p,\partial\Omega)= \inf_{q\in \partial\Omega}d(p,q),$$
where $d(p,q)$ denotes the  Euclidean distance between $p$ and $q$, and $\inf$ denotes the infimum. The signed distance field is then given by $$f_{\text{sdf}}(p)=\left\{
\begin{aligned}
    -d(p,\partial\Omega), p\in\Omega^-\\
    d(p,\partial\Omega), p\in\Omega^+
\end{aligned}
\right.$$
This function is continuous ($C^0$) and takes zero on the surface $\partial\Omega$. When $\partial\Omega$ is piecewise smooth, its gradient satisfies the eikonal equation,
$|\nabla f_{\text{sdf}}(x)|=1$,
except at points on the medial axis where the function becomes non-differentiable.
The gradient of the SDF at surface points corresponds exactly to the outward unit normal:
$$\nabla f_{\text{sdf}}(p)=N(p), p\in\partial\Omega.$$ A 2D illustration of the SDF is shown in 
Figure~\ref{fig:field2D}~(b).

\paragraph{Unsigned Distance Field}
Similar to the SDF, the UDF also encodes distances based on the Euclidean metric but omits the sign, assigning non-negative values to both interior and exterior points, as illustrated in Figure~\ref{fig:field2D}~(c).
The UDF is non-differentiable on the curved surface, at which it attains its local minimum with a value of zero. By omitting the sign, the UDF is capable of representing open surfaces and non-manifold structures, making it suitable for a broader range of geometric configurations.

\subsection{Implicit Medial Axis Representation}\label{subsec:prelimnary-MF}


\paragraph{Medial Field}
Introduced by \cite{rebain2021deep}, the \textbf{medial field} measures the local thickness of a shape.
For each point $p$, $f_{\text{mf}}(p)$ stores a scalar value representing the radius of its projected medial point  $\mathcal{M}_{\text{proj}}(p)$, defined as 
\begin{equation}
    f_{\text{mf}}(p)=f_{\text{sdf}}(\mathcal{M}_{\text{proj}}(p)).
    \label{eq:mf_value}
\end{equation}
This implies that for any medial point $q\in\mathcal{M}$, all points lying on the segment between $q$ and its closest surface point share the same MF value. Consequently, for any arbitrary point $p$, its MF value cannot be smaller than the absolute value of the SDF:
\begin{equation}
    f_{\text{mf}}{(p)}\geq |f_{\text{sdf}}{(p)}|.
    \label{eq:greater}
\end{equation}
An illustration of the MF is shown in Figure~\ref{fig:field2D}~(d).
\citet{rebain2021deep} further revealed a strong relationship between the SDF and MF: their gradients are orthogonal everywhere except on the medial axis, that is, 
\begin{equation}
    \nabla f_{\text{sdf}}(p)\cdot \nabla f_{\text{mf}}(p)=0.
    \label{eq:perpendicular}
\end{equation}
Equations~(\ref{eq:mf_value}),~(\ref{eq:greater}), and ~(\ref{eq:perpendicular}) provide the necessary and sufficient conditions for a field to qualify as a medial field. A detailed proof can be found in~\cite{rebain2021deep}.

\paragraph{Deep Medial Field}
Based on the above conditions, \citet{rebain2021deep} formulated the medial field in a variational framework and introduced three constraints for training the \textbf{deep medial field (DMF)}: 
\begin{equation}    \mathcal{L}_{\text{max}}=\int_{\mathbb{R}^3}{\max{\left(|f_{\text{sdf}}(x)|-f_{\text{mf}}(x),0\right)}^2 dx}
\end{equation}
\begin{equation}
    \mathcal{L}_{\text{ortho}}= \int_{\mathbb{R}^3}{\left(\nabla f_{\text{sdf}}(x) \cdot \nabla f_{\text{mf}}(x)\right)^2 dx}
\end{equation}
\begin{equation}\label{eq:consis}
    \mathcal{L}_{\text{consis}}= \int_{\mathbb{R}^3}{\left(|f_{\text{sdf}}(x)(\mathcal{M}_{\text{proj}}(x))|-f_{\text{mf}}(x)\right)^2 dx}
\end{equation}
The overall training objective for the medial field is then defined as
\begin{equation}
    \mathcal{L}_{\text{MF}}= \lambda_{1} \mathcal{L}_{\text{max}}+ \lambda_{2} \mathcal{L}_{\text{ortho}}
    +\lambda_{3} \mathcal{L}_{\text{consis}},
\end{equation}
where the weights $\lambda_i$ balance the three terms.

Although the MF or DMF encodes the radius information associated with the medial axis, it does not directly capture the geometric structure of the axis itself, making it difficult to extract the medial mesh from this field alone.

\section{Method}
\label{sec:method}

\begin{figure}
    \centering
    \includegraphics[width=\linewidth]{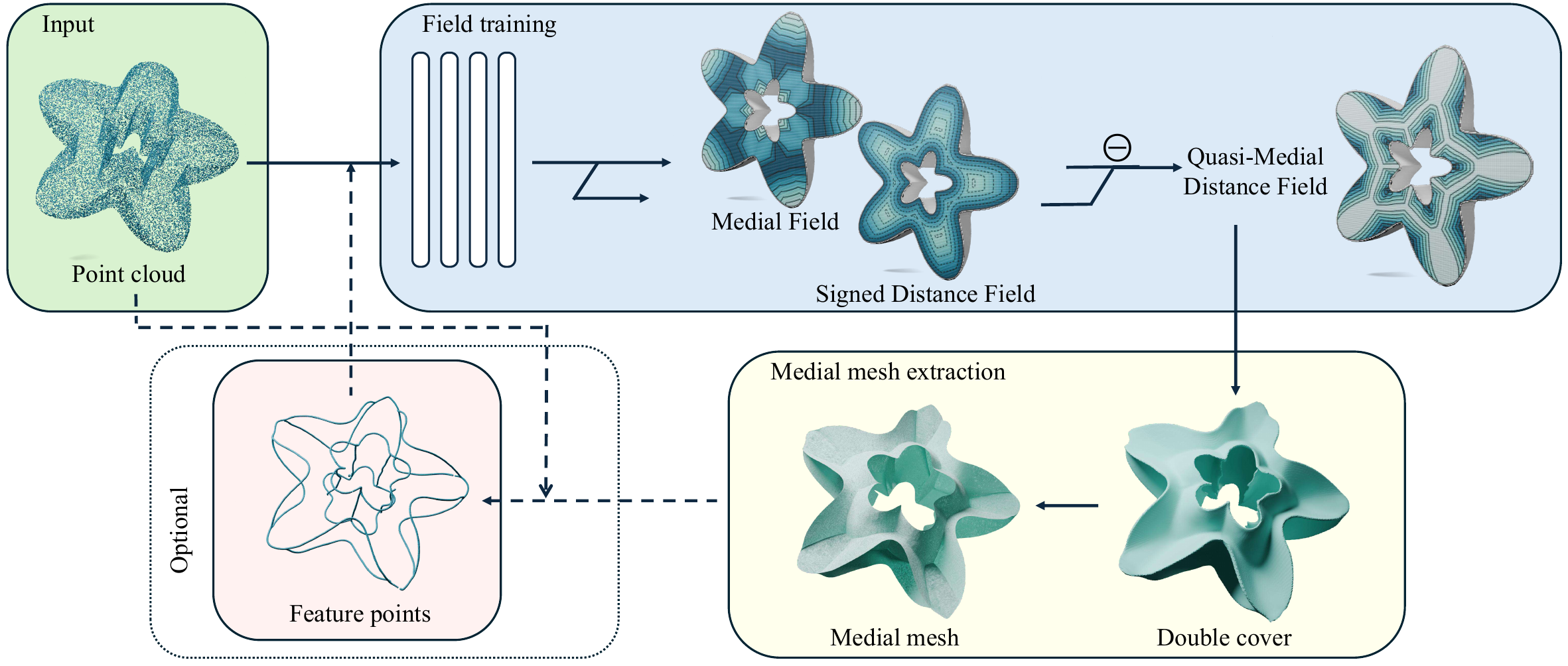}
    \caption{Pipeline of our medial axis computation method. The input consists of either a point cloud or sampled points from a given mesh. After joint training of the signed distance field and medial field, we compute the implicit medial axis representation, termed Q-MDF, by taking their difference. A double covering envelope of the medial axis is then extracted and collapsed to form a zero-volume medial membrane. Optionally, sharp features can be enhanced by detecting feature points from both the input point cloud and the medial membrane, followed by tr-training the network with this additional feature constraint.
    }
    \label{fig:pipeline}
\end{figure}

Our approach is a learning-based framework consisting of two main stages: (1) training a neural network to learn the \emph{quasi-medial distance field}, which serves as a neural implicit representation of the medial axis (Section~\ref{subsec:training}); and (2) extracting the medial axis from the learned field (Section~\ref{subsec:extract}). To further enhance the network's ability to preserve sharp features, particularly in CAD-like models, we incorporate a \emph{feature promotion strategy} (Section~\ref{subsec:feature}). An overview of the entire pipeline is illustrated in Figure~\ref{fig:pipeline}.

\subsection{Training}
\label{subsec:training}

\subsubsection{Medial Distance Fields}\label{subsubsec:mdf}
Because the medial axis $\mathcal{M}$ of a solid shape $\Omega$ is typically a non-manifold curved surface with boundaries, the unsigned distance field serves an ideal representation due to its flexibility in modeling surfaces of arbitrary topology. 
To distinguish the neural implicit representation of a solid shape from that of its medial axis, we refer to the UDF of the medial axis as the \textbf{medial distance field (MDF)}. Formally, the MDF is defined as follows:
\begin{Def}
    Given a solid shape $\Omega$ with medial axis $\mathcal{M}$, for any point $p\in\mathbb{R}^3$, let $q$ denote its closest point on $\mathcal{M}$. The medial distance field value at $p$ is defined as \[f_m(p)=\|p-q\|.\] 
\end{Def}
Figure~\ref{fig:field2D} (c) illustrates the MDF derived from a solid shape.
In fact, the MDF is continuous ($C^0$) within both $\Omega^-$ and $\Omega^+$, but it may exhibit discontinuities across the surface $\partial\Omega$, since the MDF value of any point outside the shape corresponds to its distance to the outer medial axis. As most practical applications primarily focus on the inner medial axis, we therefore restrict our subsequent analysis to $\Omega^-$ for clarity.

It is straightforward to observe that for any medial point $p\in\mathcal{M}$, the MDF value satisfies $f_m(p)=0$, and $f_m(p)$ increases monotonically as $p$ moves away from $\mathcal{M}$. 
For a sufficiently small $\epsilon>0$, the $\epsilon$-level set of $f_m$ forms a thin membrane surrounding the medial axis, which is homotopy equivalent to both the medial axis and the original shape~\cite{biasotti2008skeletal}.
Nevertheless, computing an accurate medial axis remains a difficult task, particularly for shapes with complex topology, missing regions, or noisy sampling. As a result, directly obtaining the MDF for a given shape is generally non-trivial.

\subsubsection{Quasi-Medial Distance Fields}
To facilitate MDF learning, we introduce an alternative neural implicit representation that closely resembles the MDF but is easier to compute, termed the \textbf{quasi-MDF (Q-MDF)}. This representation is motivated by the observation that the absolute SDF value increases along the gradient direction and reaches its maximum at medial points $p\in\mathcal{M}$, where $f_{\text{mf}}(p)=|f_{\text{sdf}}(p)|$ holds.
Therefore, by subtracting these two fields, we can obtain a continuous field, Q-MDF, defined as follows:
\begin{Def}
    Given a solid shape $\Omega$ with signed distance field $f_{\text{sdf}}$ and medial field $f_{\text{mf}}$, the quasi-medial distance field value at any point $p\in\mathbb{R}^3$ is defined as 
    \begin{displaymath}f_q(p) = f_{\text{mf}}(p) - |f_{\text{sdf}}(p)|.\end{displaymath}
\end{Def}

Its geometric meaning is illustrated in Figure~\ref{fig:field2D}. For an arbitrary point $p$, the Q-MDF measures the distance between $p$ and its projected medial point $\mathcal{M}_{\text{proj}}(p)$. Therefore,
Q-MDF behaves similarly to MDF: its value is zero at $\mathcal{M}$ and increases as the point moves away from the medial axis. 
However, unlike MDF, which measures the distance between $p$ and the closest point on the medial axis, the Q-MDF value at $p$ is slightly larger than that of MDF.


When extracting level sets using the same $\epsilon$ value, the membrane produced by Q-MDF is generally more compact than that of MDF,  although their overall shapes are comparable. 

In practice, this slight discrepancy is ignored.
Q-MDF serves as an effective surrogate for MDF, offering a continuous and differentiable representation that greatly simplifies the learning process.

\subsubsection{Joint Training}\label{subsec:Q-MDF}
Our network architecture is based on the neural implicit surface framework~\cite{chen2019learning,park2019deepsdf}, where a multi-layer perceptron (MLP) takes the spatial coordinates of input points as input. Building on the design of Deep Medial Field~\cite{rebain2021deep}, we employ a multi-headed architecture that jointly predicts both the SDF and the medial field.   
Specifically, the network contains two MLP heads corresponding to the inside and outside medial fields, denoted as $M^{-}(x)$ and $M^{+}(x)$, respectively, and the final output is selected according to the sign of the predicted SDF.

\subsection{Extraction}\label{subsec:extract}
Extracting the medial mesh from Q-MDF involves two main stages: (1) generating a watertight thin-sheet manifold that encloses the entire medial axis and (2) collapsing this structure to zero volume to approximate the medial surface. \modi{The final medial axis transform is then constructed from this medial membrane, with the radius at each vertex retrieved from the SDF.}

\paragraph{$\epsilon$-Covering Extraction}
Since the medial structure is represented by a UDF, it typically forms  non-manifold geometry that cannot be directly extracted using standard Marching Cubes, which is primarily designed for closed surfaces represented by SDFs. Following the optimization-based DCUDF method~\cite{hou2023robust}, we first compute an initial mesh using the conventional Marching Cubes algorithm~\cite{lorensen1998marching} to extract the 
$\epsilon$-isosurface of Q-MDF. This results in a watertight envelope that fully covers the entire medial axis and serves as the starting point for subsequent volume collapse.

\paragraph{Medial Axis Contraction}
After obtaining the $\epsilon$-isosurface of Q-MDF, the double covering is contracted to zero volume to approximate the true medial axis, which is a non-manifold curved surface. \re{To facilitate this process, we first apply tetrahedralization to the double covering, forming a closed volume that encapsulates the medial region.
We then optimize a vertex mapping $\pi$ that adjusts vertex positions while preserving the original topological connectivity by minimizing the following energy:}

\begin{align}
\min_{\pi} \ 
& \sum_{v\in\mathcal{S}_\epsilon} f_q(\pi(v)) 
+ \lambda_1 \, \text{Vol}(\mathcal{S}_\epsilon) \nonumber \\
& + \lambda_2 \sum_{v\in\mathcal{S}_\epsilon} 
w(v) {\left\|\pi(v) - \frac{1}{\mathcal{N}(v)} 
\sum_{v_i\in\mathcal{N}(v)} \pi(v_i)\right\|}^2,
\end{align}
where $\mathcal{S}_\epsilon$ denotes the $\epsilon$-isosurface and $\text{Vol}(\mathcal{S}_\epsilon)$ measures the interior volume it encloses. The first term $f_q(\pi(v))$ encourages 
vertices to move toward local minima of the learned Q-MDF. The second term promotes volume shrinkage toward zero, driving the surface to converge more rapidly to the true medial axis. The third term acts as a surface Laplacian regularizer that  enforces local smoothness of the contracted geometry. The vertex weight $w(v)$ is inversely proportional to the total area of the triangles incident to $v$, and $\mathcal{N}(v)$ denotes the one-ring neighborhood of $v$.

\paragraph{Remark} While the Q-MDF is theoretically zero along the medial axis when derived from ground-truth SDF and MF, the learned field may exhibit small deviations due to numerical errors or model underfitting. Crucially, these deviations are typically negligible, and the extracted medial axis does not depend on Q-MDF attaining an exact zero; instead, it corresponds to the local minima of the field, which preserve the \textbf{centering property} of the axis. In particular, the local minima of Q-MDF align with those of the SDF along the medial axis, ensuring that the extracted axis remains properly centered and geometrically consistent. To robustly extract the medial axis under these conditions, we adopt a moderate $\epsilon$ threshold to generate an $\epsilon$-isosurface around the medial axis, which empirically provides a good balance between suppressing numerical noise and preserving the continuity and topological structure of the true axis. Section~\ref{sec:5.2} provides experimental validation of these properties.

\subsection{Feature Enhancement}\label{subsec:feature}
For shapes with sharp features, the mathematically defined medial axis should extend to the feature lines, where the corresponding radius becomes zero~\cite{wang2022computing}. However, the Q-MDF obtained from joint training tends to be overly smooth if no feature priors are incorporated, causing the resulting medial axis to deviate from these feature lines. To address this, we introduce a sharp feature enhancement module, which serves as an optional component to improve both the accuracy of the medial axis and the fidelity of the field representation around sharp features. Section~\ref{subsubsec:detection} describes how sharp feature points are inferred from the input point cloud using the existing medial mesh, while Section~\ref{subsubsection:training_feature} discusses how these feature point priors are incorporated to enhance the joint training of SDF and MF.

\subsubsection{Medial Axis-Guided Feature Detection}
\label{subsubsec:detection}
The medial axes of 3D shapes are typically non-manifold curved surfaces whose boundaries consist of points with neighborhoods homeomorphic to half-disks. In a discrete mesh representation, such boundaries can be identified by detecting edges adjacent to only one face. 
However, the medial mesh extracted by our method is watertight and manifold, preventing direct boundary extraction through simple face-adjacency checks.

Due to the zero-volume nature of our medial mesh, sharp turns appear at the boundaries, which can be identified by examining dihedral angles. Specifically, if the dihedral angle of an edge is close to $2\pi$, the edge is regarded as a boundary edge. For shapes with sharp features, the medial axis extends toward these regions, causing the boundary of the medial axis to coincide with the sharp features of the shape. However, when the learned MF and SDF are overly smoothed near sharp features, the extracted medial axis may not fully reach the feature lines. Nevertheless, the boundary of the medial axis still indicates the direction of these latent features. Therefore, if an edge exhibits both low MF and SDF values, it can be extended toward the corresponding sharp feature.

\begin{figure}[htb]
    \centering
    \begin{overpic}[scale=0.4,unit=1mm]{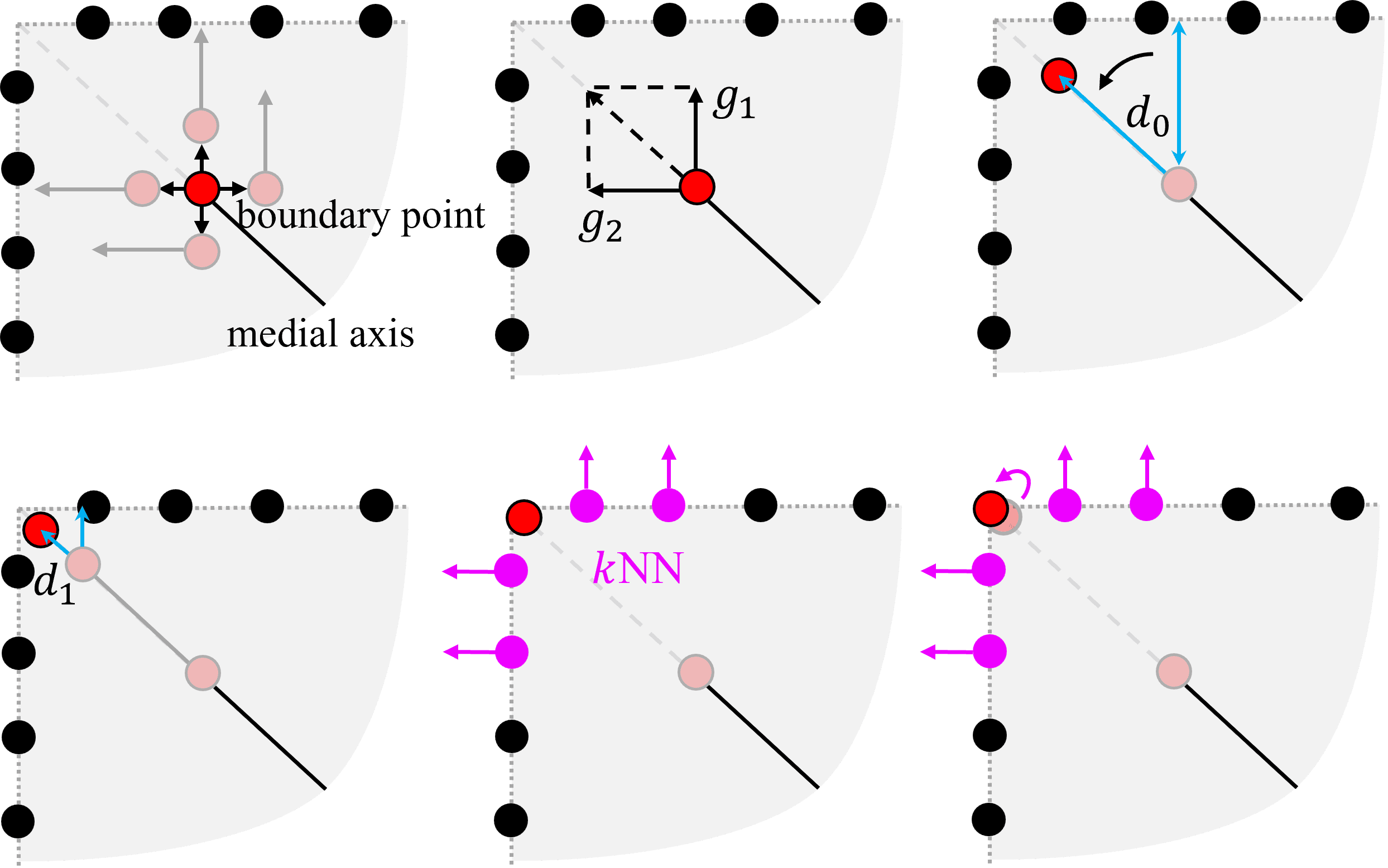}
    \put(15,33){(a)}
    \put(50,33){(b)}
    \put(82,33){(c)}
    \put(15,-2){(d)}
    \put(50,-2){(e)}
    \put(82,-2){(f)}
    \end{overpic}
    \caption{\re{2D illustration of feature point localization. (a) Sample gradients at four neighboring points. (b) Cluster them into two directions representing bilateral gradients; the marching direction is determined by their sum at the boundary point. (c)-(d) March along this direction while gradually decreasing the step size $d_i$ until it approaches zero. (e) Identify the $k$ nearest surface points to the shifted boundary point. (f) Apply a quadratic error metric to determine the feature point  position for this corner.}}
    \label{fig:FeaturePointCal}
\end{figure}

After identifying the medial axis boundary that points to potential sharp features, we uniformly sample a set of points on it and use a marching procedure to infer the corresponding feature points. 
Figure~\ref{fig:FeaturePointCal} provides a 2D illustration of this process.
\re{For each boundary point $p$ where marching is required, we first sample six neighboring positions obtained by offsetting $p$ along the positive and negative coordinate axes and query their SDF gradients. These gradients are grouped into two angularly distinct clusters, representing two dominant local directions $g_1(p)$ and $g_2(p)$. The marching direction is determined by the vector sum $g_1(p)+g_2(p)$.}
\re{We then advance the point iteratively along this direction, using the local SDF values to guide the step size, until the sampled SDF approaches zero.}
The initially inferred feature point may not be perfectly accurate or consistent with neighboring  points. To refine it, we identify the $k$ nearest surface points to the inferred location and adjust its position using the quadric error metric (QEM)~\cite{garland1997surface}. \re{Following~\cite{gelas2009variational}, we use the positions and normals of neighboring points, where the SDF gradient serves as the normal direction, to project the point onto the feature line, as illustrated in Figure~\ref{fig:FeaturePointCal}~(f).}

\subsubsection{Training with Feature Prior}
\label{subsubsection:training_feature} The set of prominent feature points $P$, either inferred from the point cloud or computed from the input mesh using a user-defined dihedral angle threshold, can be utilized to enhance the joint training of SDF and MF through additional loss terms:
$$\mathcal{L}_{\text{sharp1}}=\sum_{i=1}^n{f_{\text{sdf}}(p_i)},$$
and
$$\mathcal{L}_{\text{sharp2}}=\sum_{i=1}^n{f_{\text{mf}}(p_i)}.$$ These two losses enforce zero-valued constraints on both the SDF and MF at feature point locations, guiding the learned fields to accurately capture sharp edges and corners.

\section{Evaluation and Comparison}
\subsection{Setup}
\paragraph{Implementation Details}
All tested 3D models are normalized to fit within a unit box centered at the origin. For point cloud inputs, the raw points are directly fed into the neural network, while, for mesh inputs, we uniformly sample points on the surface before training. Both input types are processed in batches of 4,096 points, and no additional normal information is required. The joint SDF-MF network follows the architecture proposed by~\cite{rebain2021deep}. Detailed parameter settings for extracting the double covering from Q-MDF are discussed in Section~\ref{sec:5.2.3}. All experiments were performed on an NVIDIA 4090 GPU with 24 GB of memory.

\paragraph{Evaluation Metrics}
Following commonly used criteria for medial axis evaluation, we assess the  reconstruction accuracy of the medial axis transform. In theory, the reconstructed shape corresponds to the swept volume of spheres with varying radii centered along the medial axis. 
As discussed by~\cite{li2015q}, this reconstructed shape can be computed as the union of the enveloping volumes of all  medial primitives. Reconstruction accuracy is then quantified using two distance-based metrics: the two-sided mean distance error, measured by the \textbf{Chamfer distance (CD)}, and the two-sided maximum distance error, measured by \textbf{Hausdorff distance (HD)}.

We also evaluate the topological correctness of the computed medial axes using the \textbf{Euler characteristic} ($\chi$), defined as $\chi = V - E + F - C$, where $V$, $E$, $F$, and $C$ denote the numbers of vertices, edges, faces, and volumetric cells (tetrahedra in our setting), respectively.

Finally, we use \textbf{boundary smoothness} to evaluate the compactness of the medial axis. Boundary smoothness is measured by the average curvature $C_{\text{avg}}$, formulated as:
$$C_{\text{avg}}=\frac{1}{L}\int_0^{L}{\kappa(s)}ds,$$
where $\kappa(s)$ denotes the curvature and $L$ is the total curve length. For a discrete curve $C$ composed of sequential points, $C_{\text{avg}}$ can be approximated as:
$$C_{\text{avg}}=\frac{1}{L}\sum_{i=1}^{n}{l_i\theta_i},$$
where 
\[l_i = \frac{\|c_{i-1}-c_{i}\|+\|c_{i+1}-c_{i}\|}{2},\]
\[\theta_i=\pi-\arccos\left(\frac{(c_{i-1}-c_{i})\cdot(c_{i+1}-c_{i})}{\|c_{i-1}-c_{i}\|\|c_{i+1}-c_{i}\|}\right),
\]
and $n$ is the number of turning angles $\theta_i$.

\subsection{Experimental Results}\label{sec:5.2}

\subsubsection{Centerness}
\begin{figure}[htb]
    \centering
    \includegraphics[width=\linewidth]{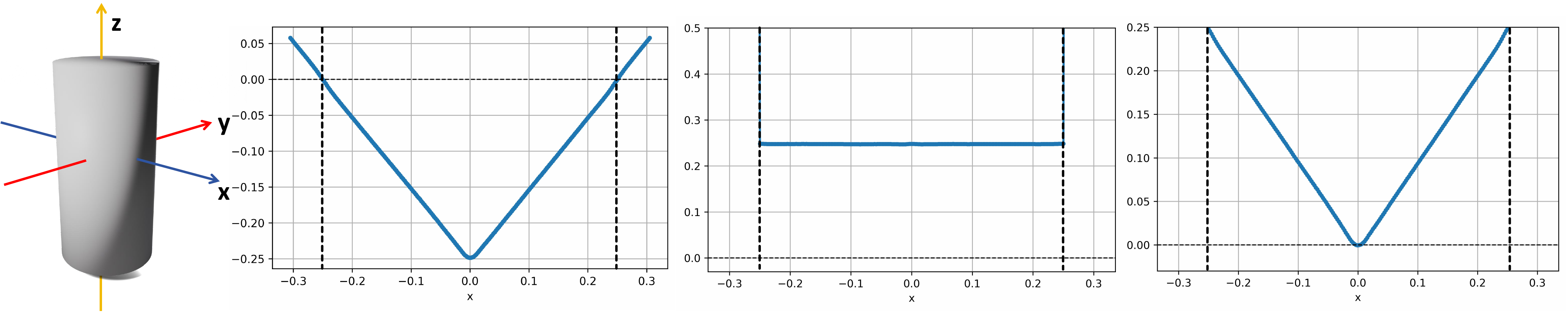}
     \makebox[0.15\linewidth][c]{Input}
    \makebox[0.27\linewidth][c]{SDF }
    \makebox[0.27\linewidth][c]{MF }
    \makebox[0.27\linewidth][c]{Q-MDF }
    \caption{\re{SDF, MF, Q-MDF values queried along the $x$-axis of a cone, showing that the local minimum of the deep Q-MDF occurs at the central axis ($x=0$).
    }}
    \label{fig:SDF-line}
\end{figure}

\begin{figure}[htb]
    \centering
    \includegraphics[width=\linewidth]{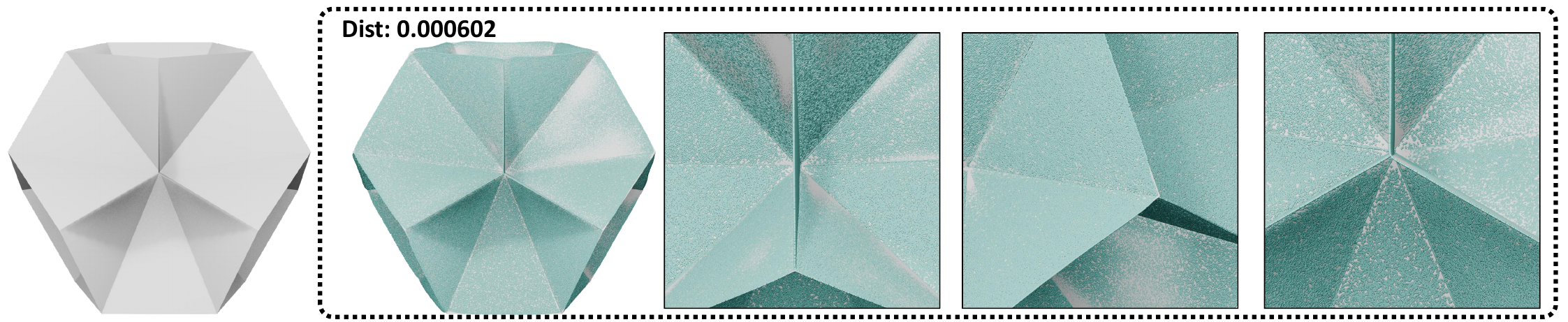}
    \makebox[0.2\linewidth][c]{GT MA}
    \makebox[0.78\linewidth][c]{Points Mapped via Q-MDF }
    \caption{\re{One million points are uniformly sampled in $\mathbb{R}^3$ and mapped to the local minima of the deep Q-MDF. The mapped points closely align with the ground-truth medial axis (GT MA), with a mean distance of \(6.02\times 10^{-4}\), further confirming the centerness of our deep Q-MDF.}
    }
    \label{fig:sampling}
\end{figure}

\re{To ensure the centerness property of the extracted medial axis, the local minima of the learned Q-MDF should coincide with true medial axis. We verify this property by analyzing the behavior of the SDF, MF, and Q-MDF along the $x$-axis of a cylindrical shape (Figure~\ref{fig:SDF-line}). 
As shown in the plot, the Q-MDF reaches its local minima at the same positions where the SDF attains its minimum, corresponding to the geometric center of the shape. This observation confirms that the centerness of the Q-MDF is governed by the alignment between the local minima of the SDF and the true medial axis. To further validate this property in 3D space, we randomly sample spatial points and project them onto their local minima by following the negative gradient of the Q-MDF. As illustrated in Figure~\ref{fig:sampling}, the projected points closely align with the ground-truth medial axis, demonstrating that the local minima of our learned Q-MDF exhibit strong centerness and geometric consistency.}


\subsubsection{Accuracy}\label{subsubsec:accuracy}
\begin{figure}[htb]
    \centering
    \begin{minipage}{0.9\linewidth}
    \includegraphics[width=0.24\linewidth,trim=1cm 1cm 4cm 1cm, clip]{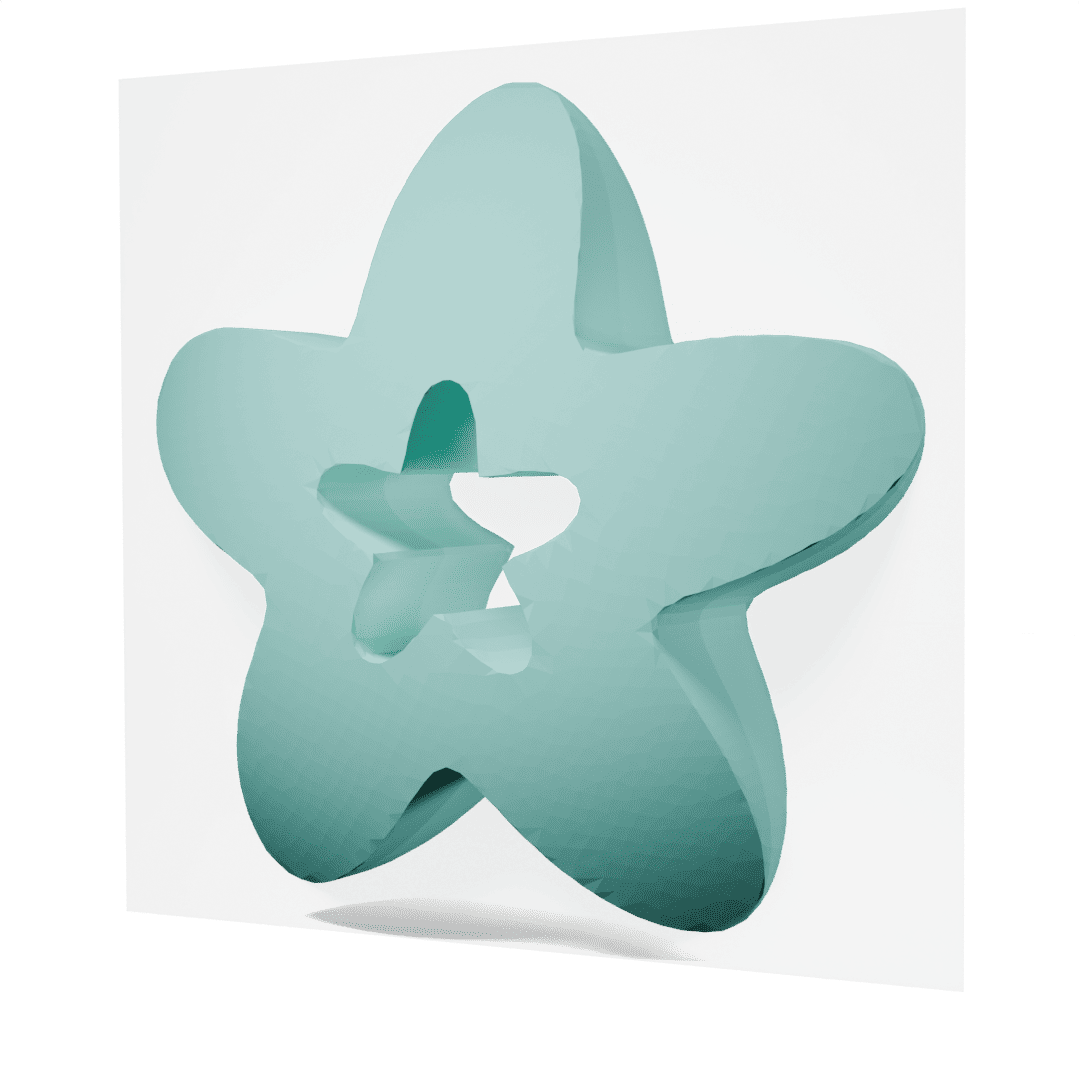}
    \includegraphics[width=0.24\linewidth]{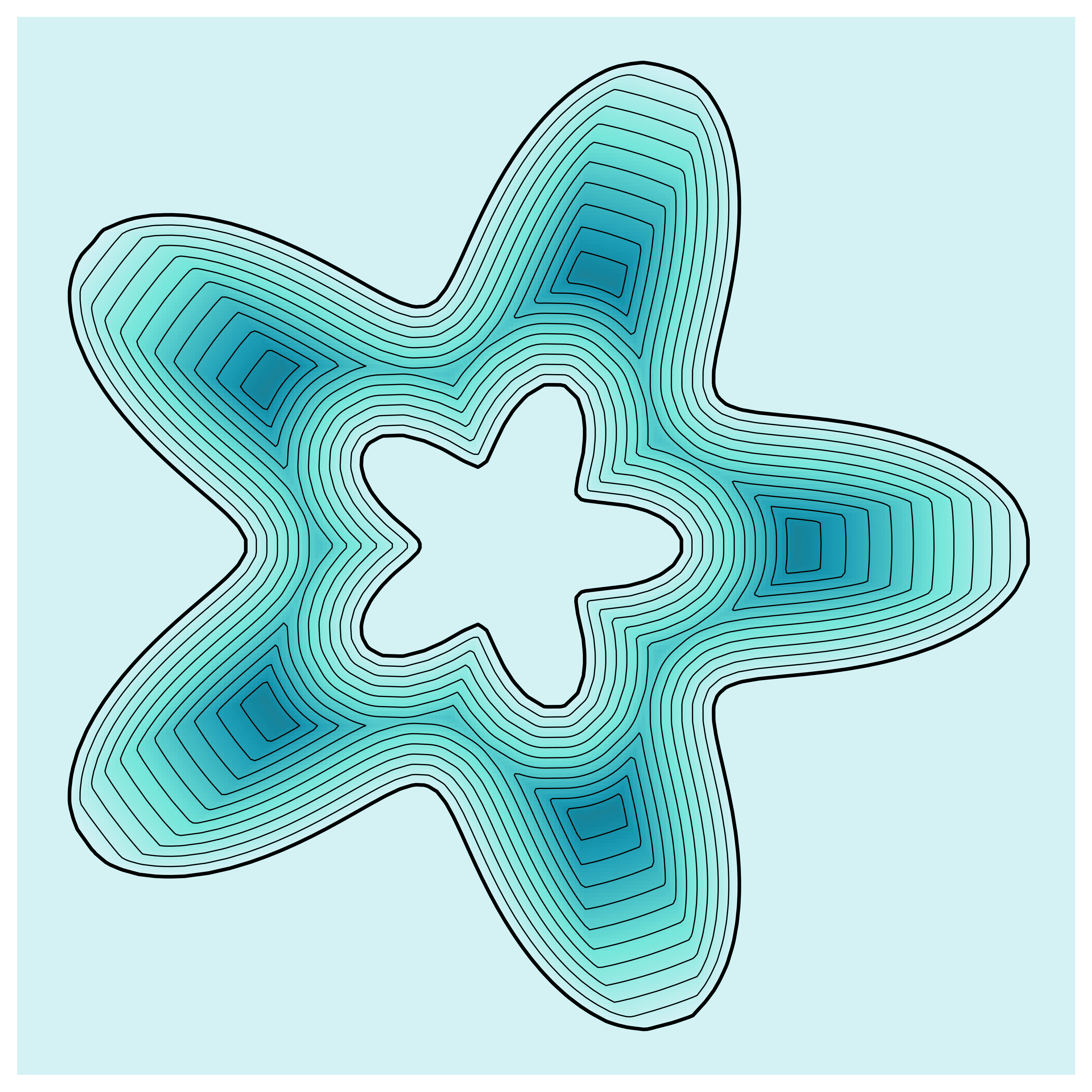}
    \includegraphics[width=0.24\linewidth]{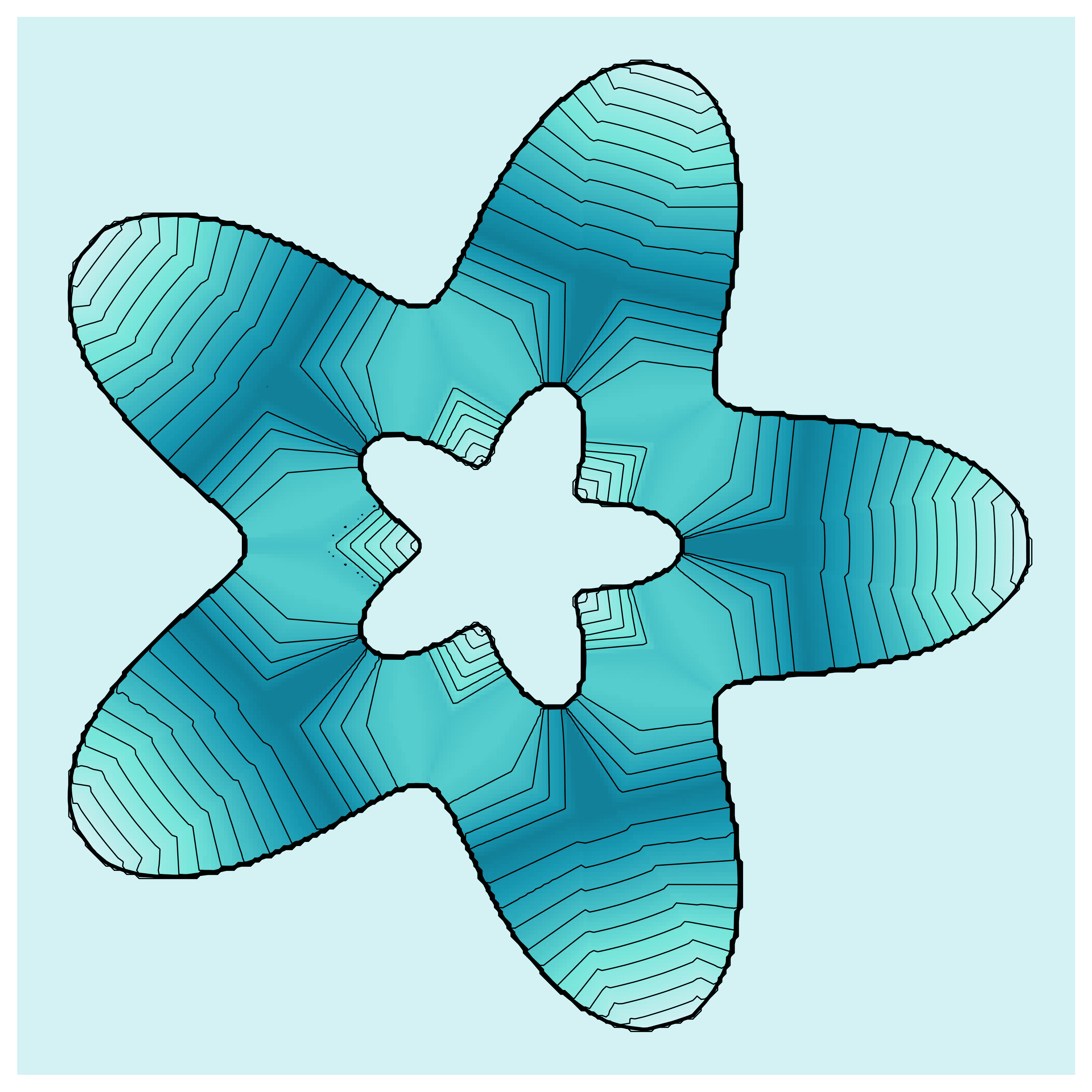}
    \includegraphics[width=0.24\linewidth]{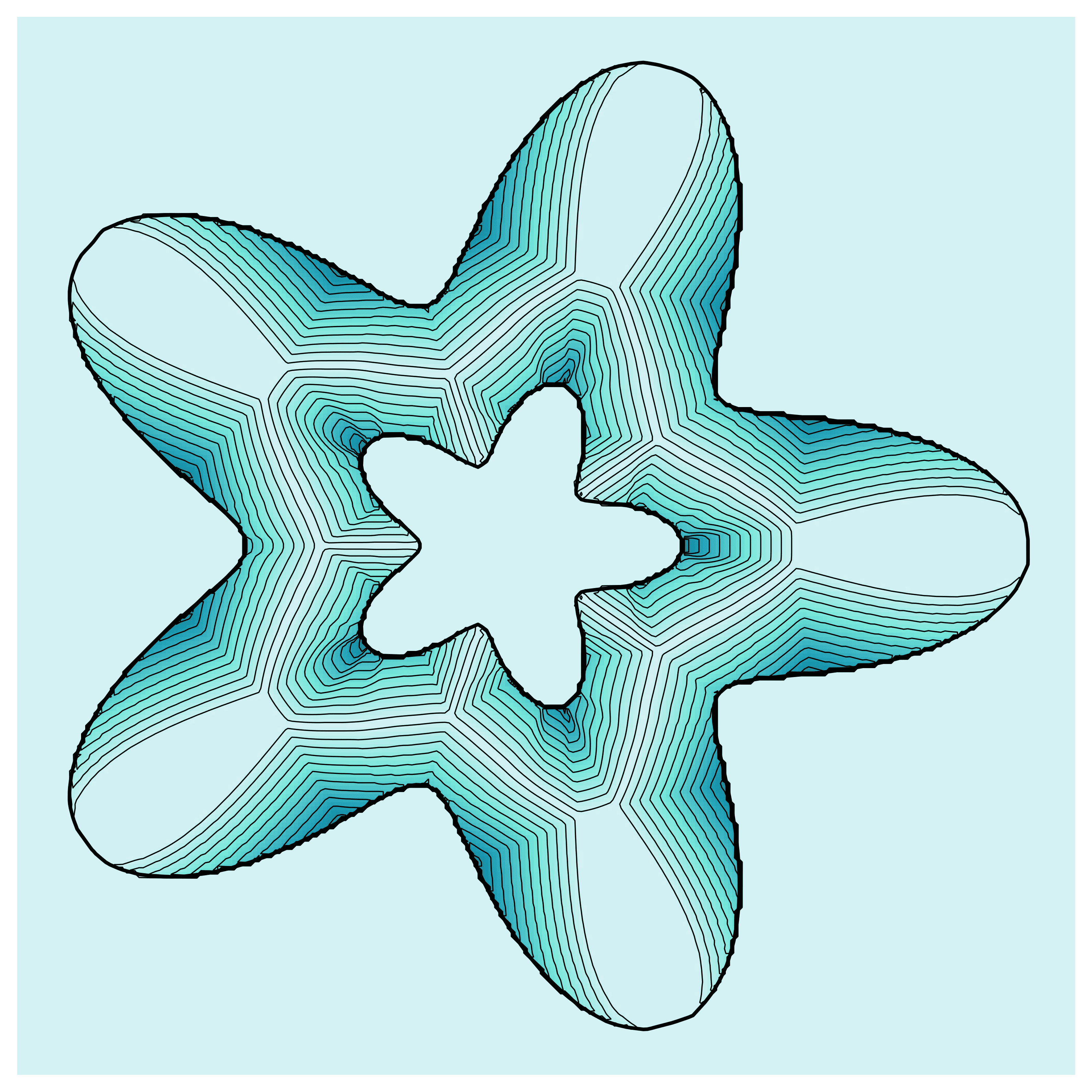}
    \makebox[0.24\linewidth][c]{(a) Input}
    \makebox[0.24\linewidth][c]{(b) SDF}
    \makebox[0.24\linewidth][c]{(c) MF}
    \makebox[0.24\linewidth][c]{(d) Q-MDF}
    \includegraphics[width=0.24\linewidth,trim=1cm 1cm 1cm 1cm, clip]{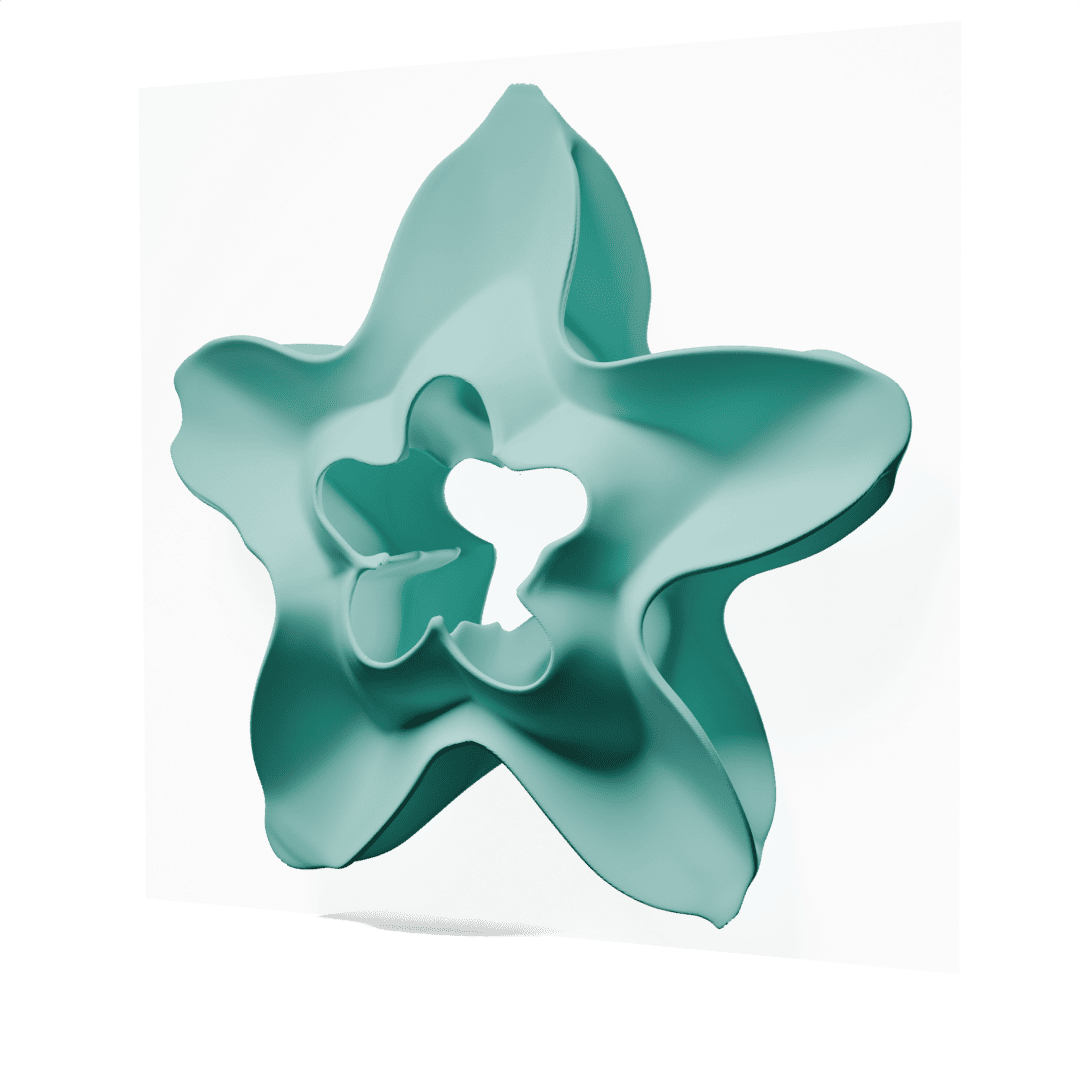}
    \includegraphics[width=0.24\linewidth]{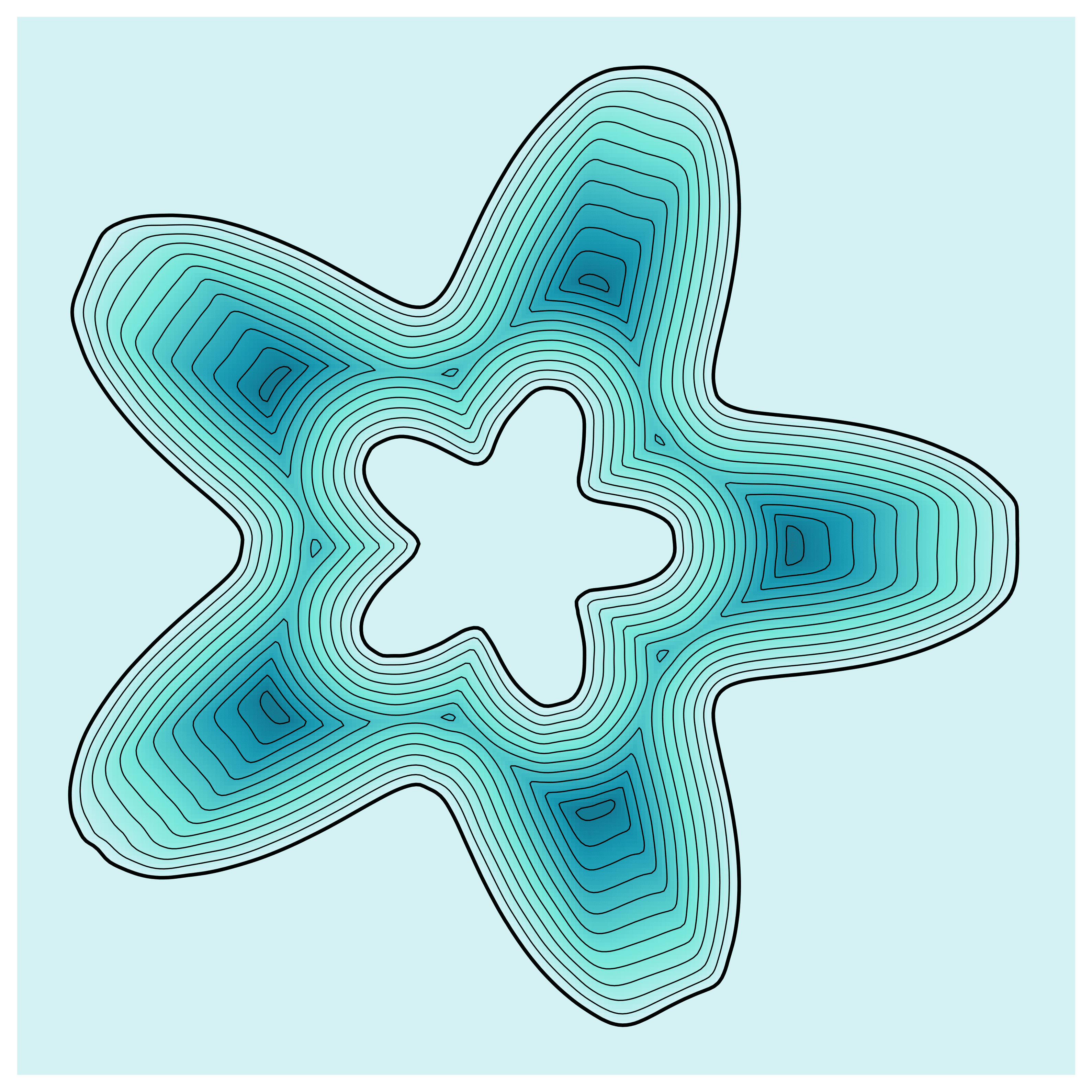}
    \includegraphics[width=0.24\linewidth]{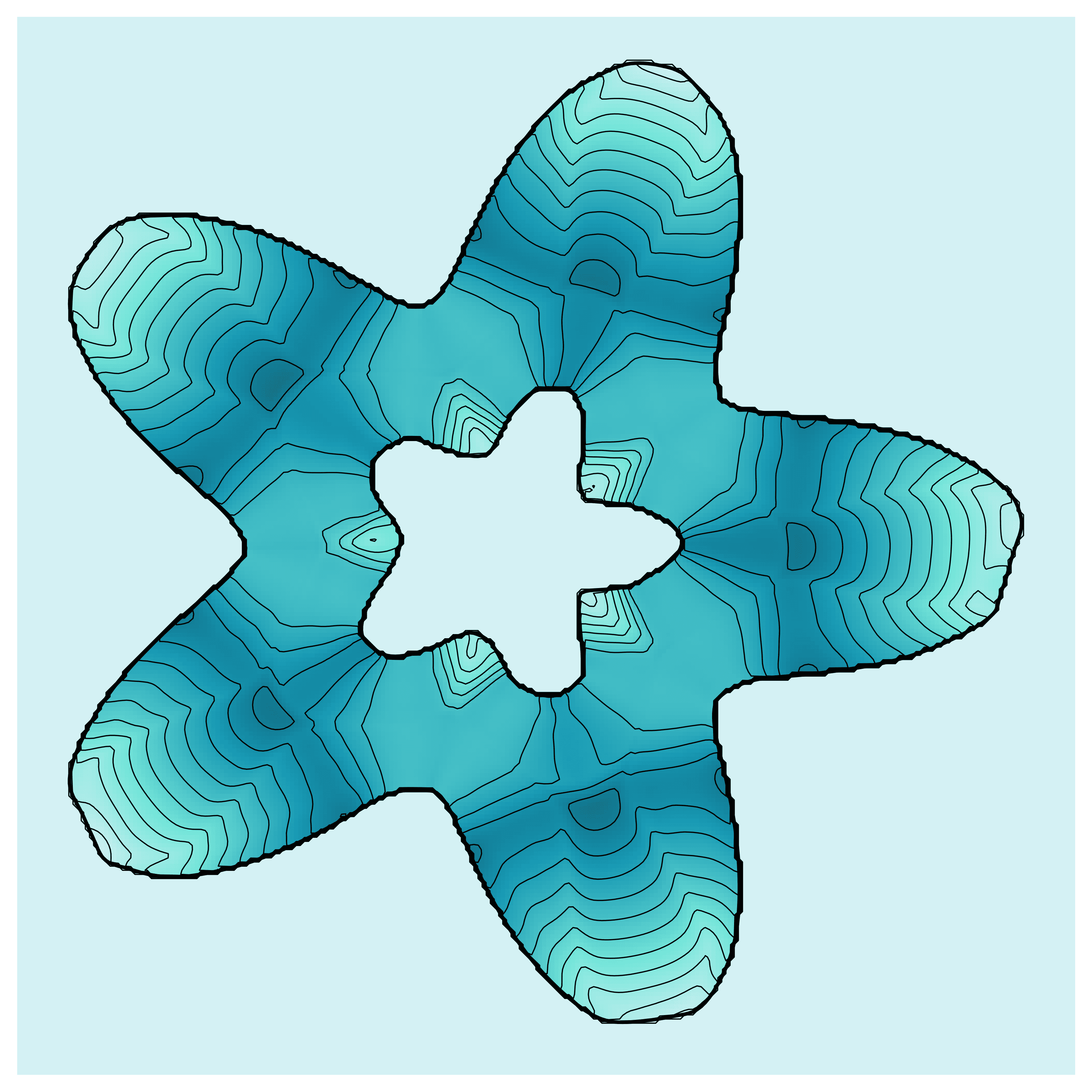}
    \includegraphics[width=0.24\linewidth]{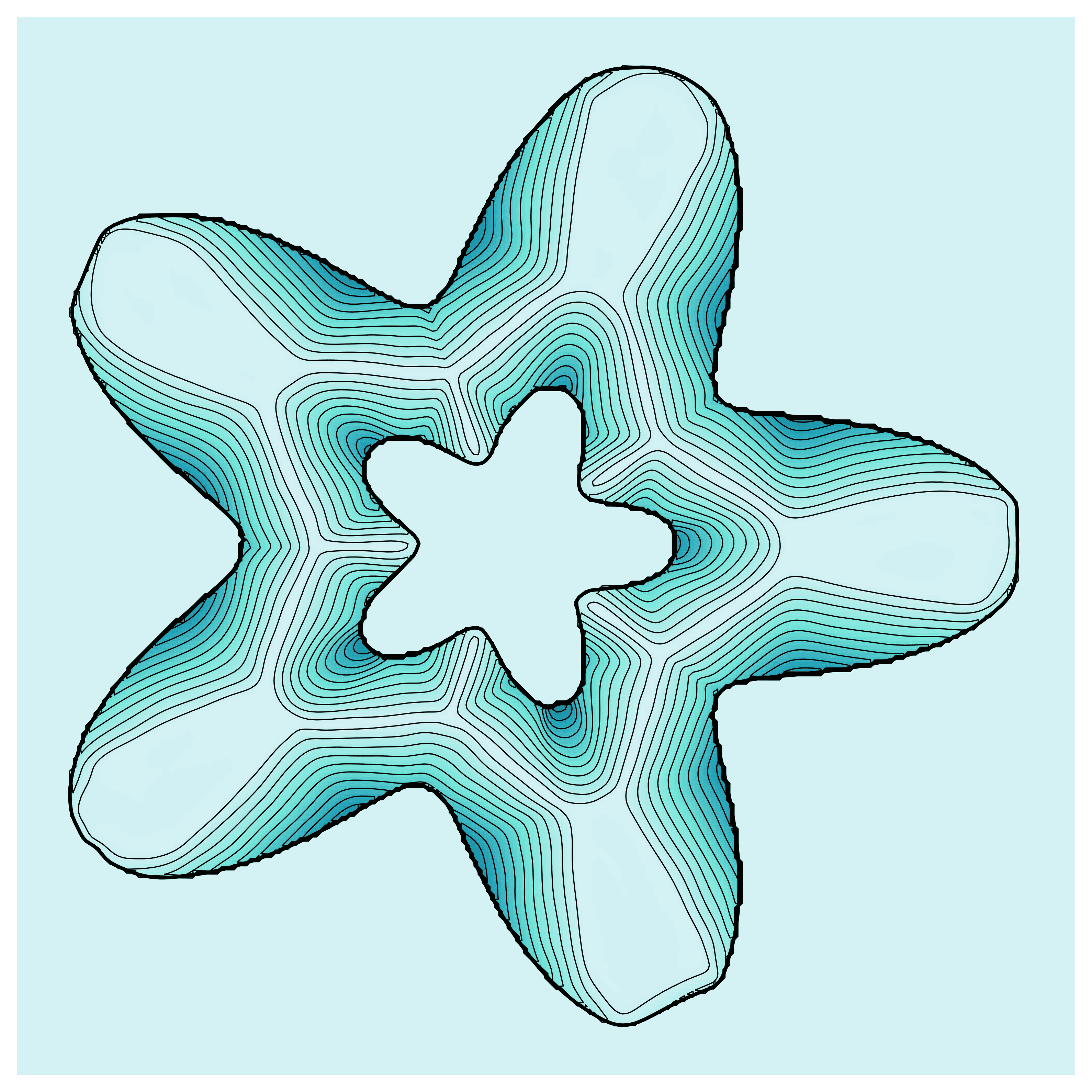}
    \makebox[0.24\linewidth][c]{(e) DC}
    \makebox[0.24\linewidth][c]{(f) SDF (GT)}
    \makebox[0.24\linewidth][c]{(g) MF (GT)}
    \makebox[0.24\linewidth][c]{(h) Q-MDF (GT)}
     \end{minipage}
    \hfill
    \begin{minipage}{0.08\linewidth}
        \centering
       \includegraphics[height=4.5cm]{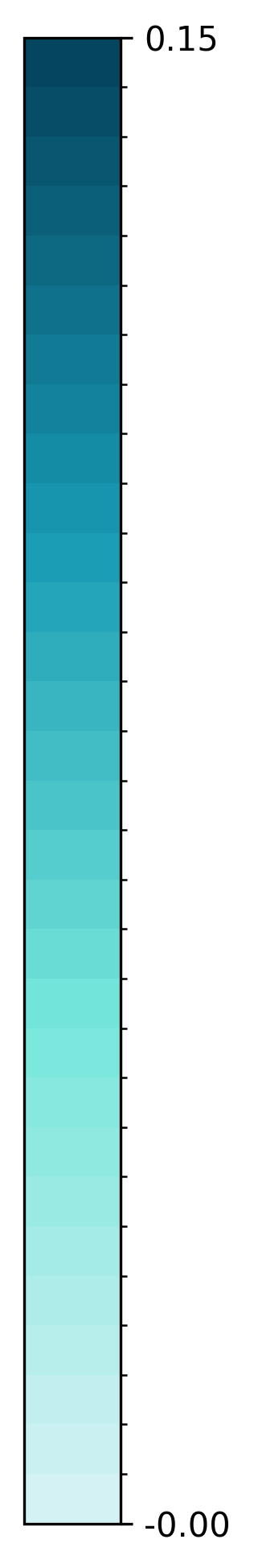}
    \end{minipage}
    \caption{Accuracy analysis. (a) A star-shaped 3D object of genus 1. (b)-(d) Visualizations of the SDF, MF and Q-MDF on a cut plane. (e) Initial double covering extracted from our Q-MDF with $\epsilon=0.005$. (f)-(h) Corresponding ground-truth fields.
    }
    \label{fig:fields_accuracy}
\end{figure}

\begin{figure}[htb]
    \centering
    \includegraphics[width=1\linewidth]{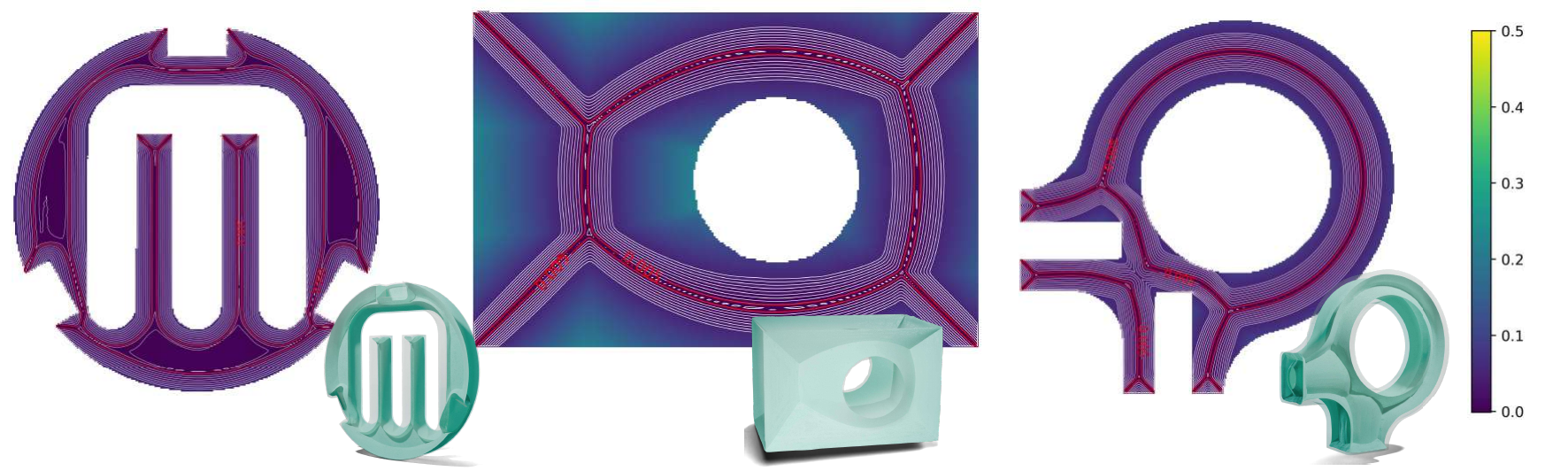}
    \caption{\re{Cross-sections of the learned Q-MDF. For each model, ten level sets with isovalues ranging from 0 to 0.05 (white) are visualized, along with the specific contour at $\epsilon=0.005$ (red). 
The $\epsilon$-level set forms a compact envelope enclosing the entire medial structure.}}
    \label{fig:covering}
\end{figure}

The learned Q-MDF obtained from joint training closely approximates the ground-truth Q-MDF. Since the loss terms are applied over the entire spatial domain, the resulting field satisfies the variational constraints not only near the surface but also in regions farther away. 
Figure~\ref{fig:fields_accuracy} compares \re{our predicted fields with the ground truth, showing that the learned Q-MDF exhibits a strong resemblance of the true field.}
Therefore, the resulting Q-MDF values around the medial axis are highly reliable, allowing a small threshold $\epsilon$ to be used for extracting the initial \re{double covering} that tightly wraps the medial axis.
\re{As shown in Figure~\ref{fig:covering}, when $\epsilon=0.005$, the corresponding $\epsilon$-isosurface (outlined in red) forms a compact envelope that fully wraps the medial axis, preserving its overall geometry and local structure. Although the resulting surface is a double-covered mesh rather than a single non-manifold sheet, it remains a faithful representation of the underlying medial structure.}

\subsubsection{Parameter Setting of $\epsilon$}\label{sec:5.2.3}
\begin{figure}[htb]
    \centering
    \includegraphics[width=\linewidth]{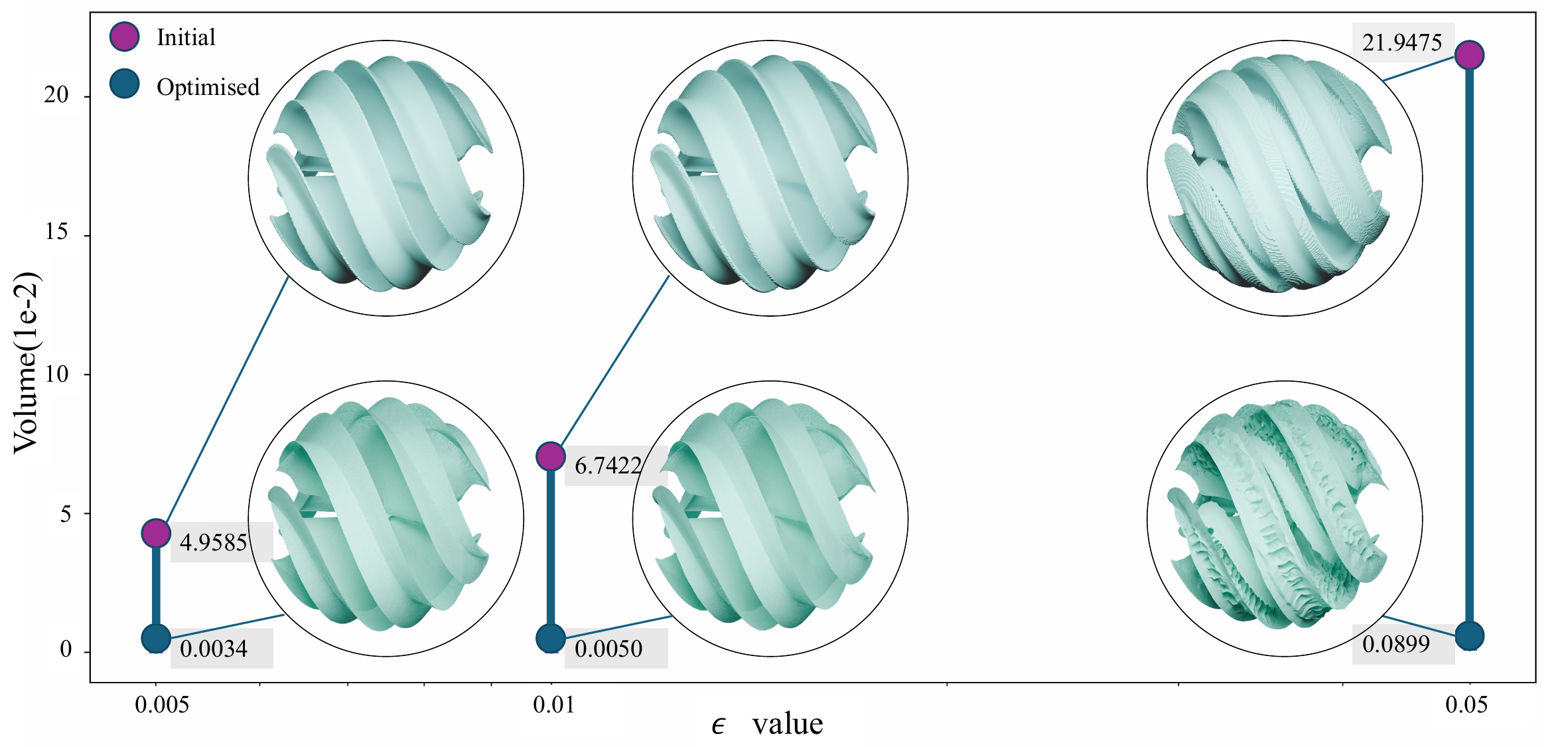}
    \caption{Visualization of shrink optimization results with different $\epsilon$ values ($0.05$, $0.01$, and $0.005$) after 3k iterations. The initial and final volumes are reported for each case. 
    }
    \label{fig:sharp_sphere_r}
\end{figure}

The zero-volume constraint remains  effective across a wide range of $\epsilon$ values.
As shown in Figure~\ref{fig:sharp_sphere_r}, increasing $\epsilon$ enlarges the volume of the initially extracted double covering. When $\epsilon$ is small, the double covering aligns more closely with the true medial axis, leading to faster and more accurate contraction during optimization. In contrast, if $\epsilon$ is too large, the shell becomes excessively thick, and redundant branches may appear after shrinkage. Nevertheless, despite these variations in initial volume, all cases eventually converge to near-zero volume after 3k iterations of shrink optimization.
\re{In this paper, we empirically set $\epsilon = 0.005$, which provides a good balance between stability and precision, yielding an isosurface that tightly encloses the medial axis, and conduct medial axis extraction experiments under this setting.}

\subsubsection{Localized Symmetric Pattern} 
\begin{figure}[htb]
    \centering
    \includegraphics[width=0.23\linewidth]{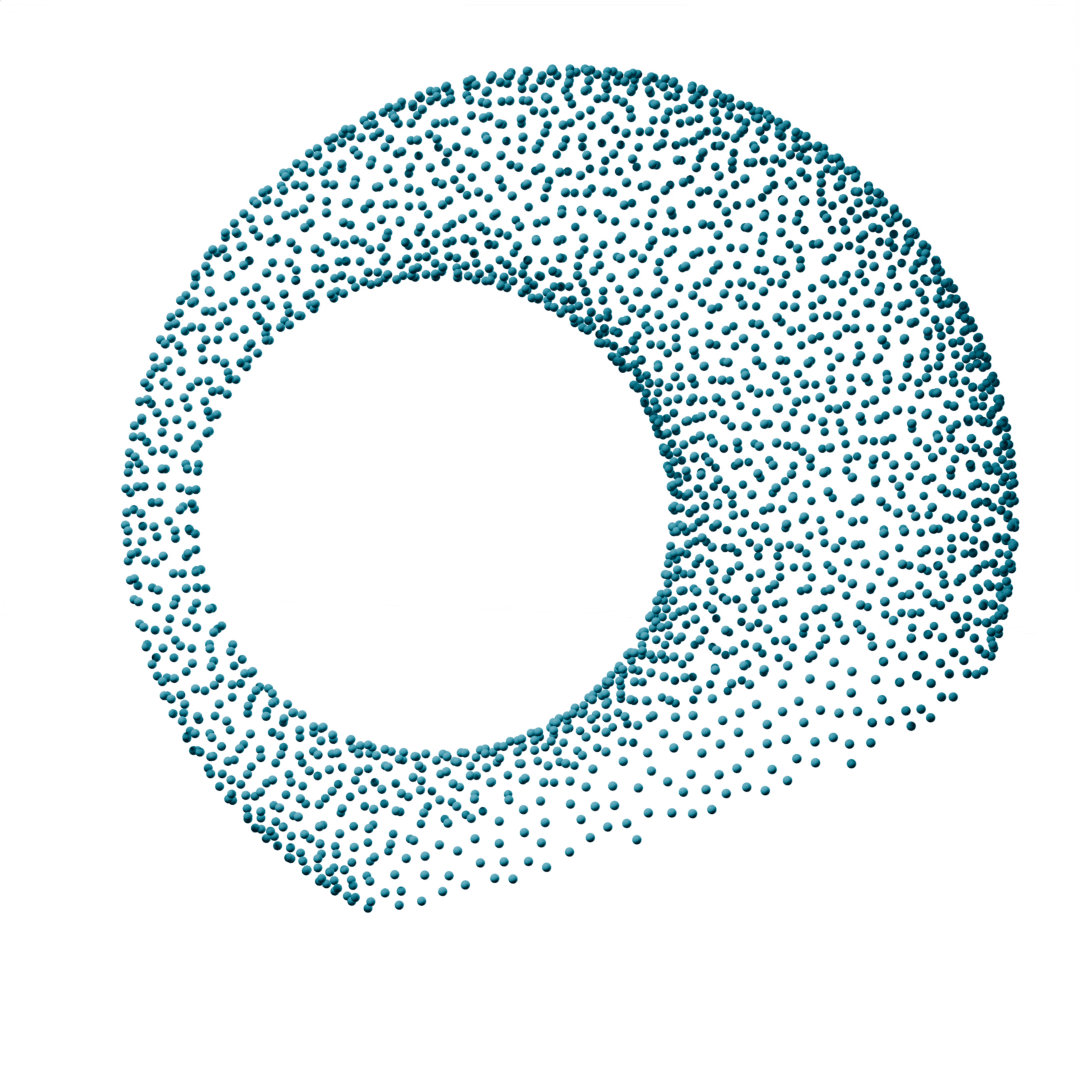}
    \includegraphics[width=0.23\linewidth]{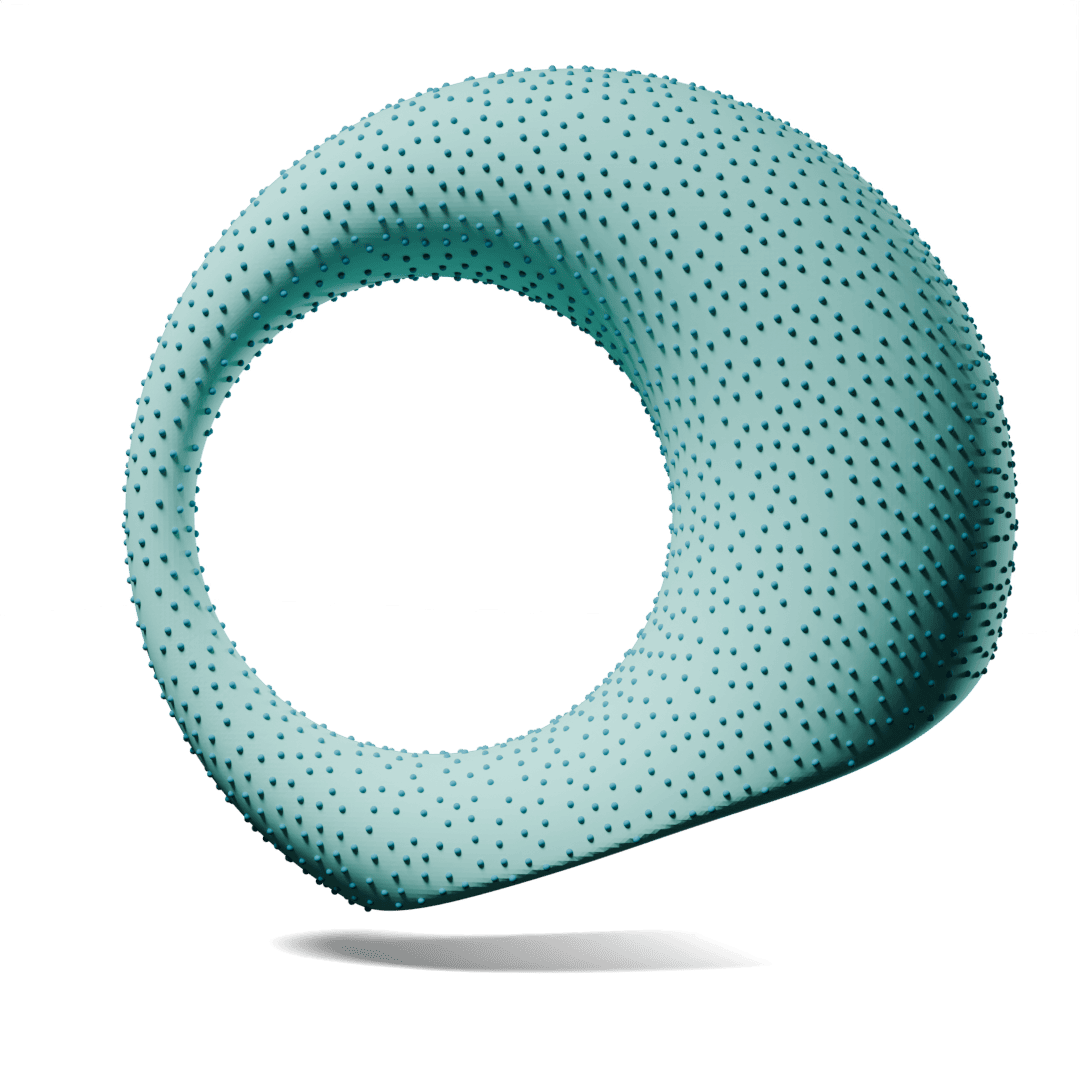}
    \includegraphics[width=0.23\linewidth]{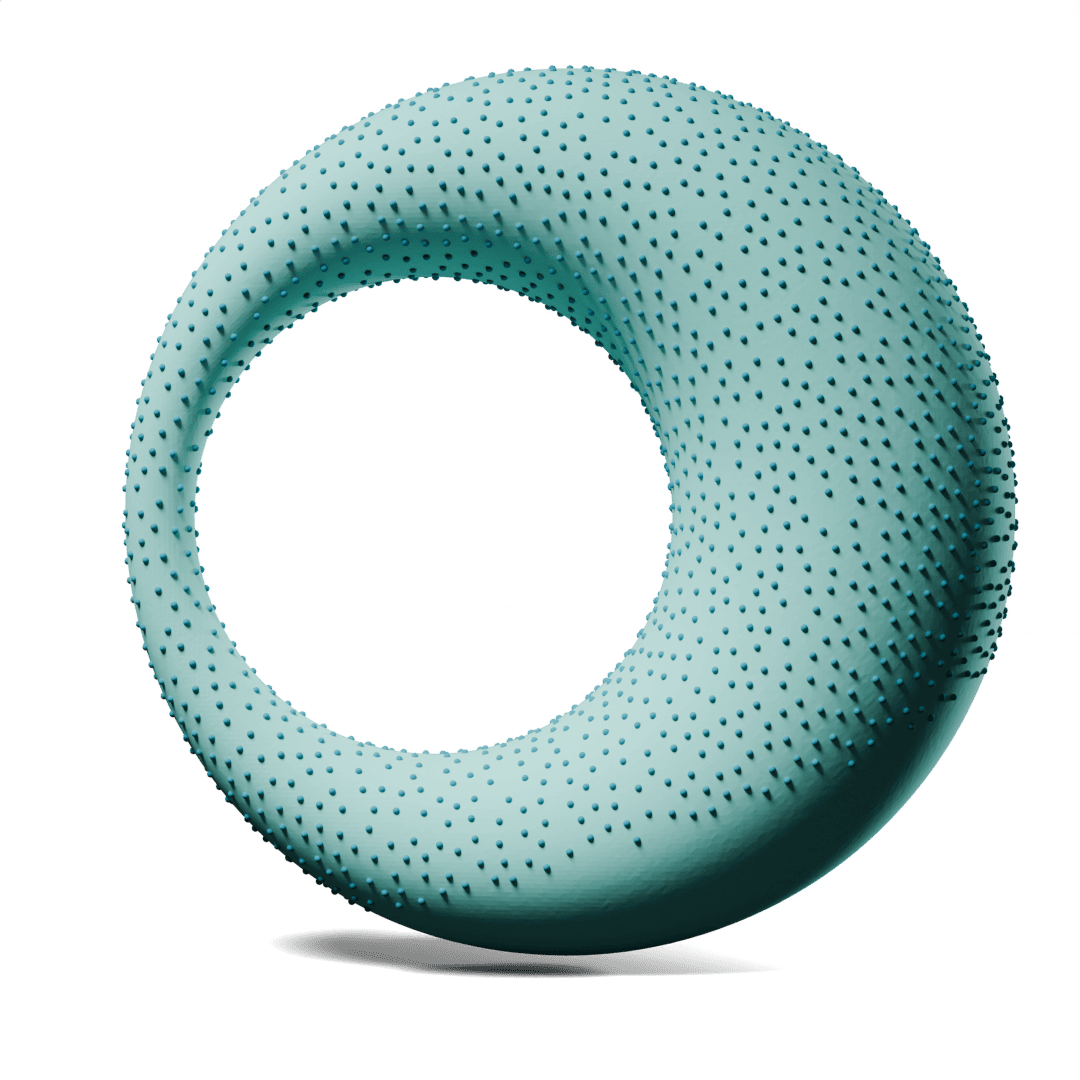}
    \includegraphics[width=0.23\linewidth]{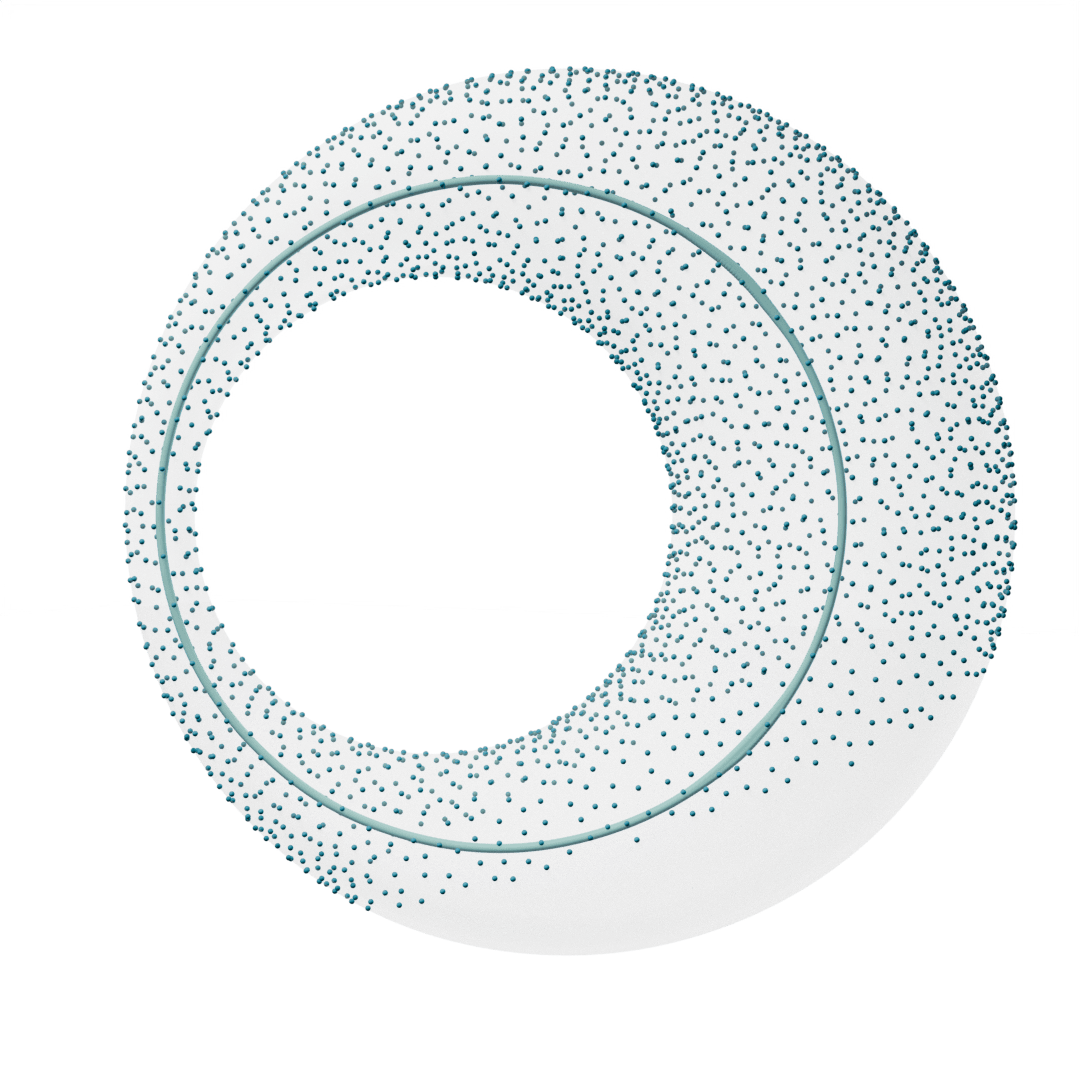}
    \makebox[0.23\linewidth][c]{(a) Input}
    \makebox[0.23\linewidth][c]{(b) w/o MF}
    \makebox[0.23\linewidth][c]{(c) w/ MF}
    \makebox[0.23\linewidth][c]{(d) Medial axis}
    \caption{Comparison of results with and without MF constraints. (a) Partially missing Dupin cyclide point cloud used as input.  
    (b)–(c) Extracted zero-isosurfaces from the corresponding SDFs. 
    (d) Medial axis computed from the fields with MF constraints. MF-guided training more effectively reconstructs the latent shape with higher smoothness and yields a more compact medial axis.}   
    
    \label{fig:symmetry}
\end{figure}

A useful property of MF-guided training is that the resulting neural shape representation tends to recover latent shapes exhibiting localized symmetry, producing smoother surfaces. As illustrated in Figure~\ref{fig:symmetry}, 
the input is a partially cut Dupin cyclide point cloud. When trained using only the SDF loss, the model fills the missing region by forming a planar patch over the gap. In contrast, with localized symmetric guidance from the MF constraints, the model infers the missing areas by smoothly ``inflating'' them, closely approximating the original geometry. Consequently, the extracted medial axis is both smooth and compact.

\begin{figure}[htb]
    \centering
     \raisebox{0.3\height}{\makebox[0.01\textwidth]{\rotatebox{90}{\makecell{\small After \hspace{1.8em} Before}}}}
    \includegraphics[width=0.95\linewidth]{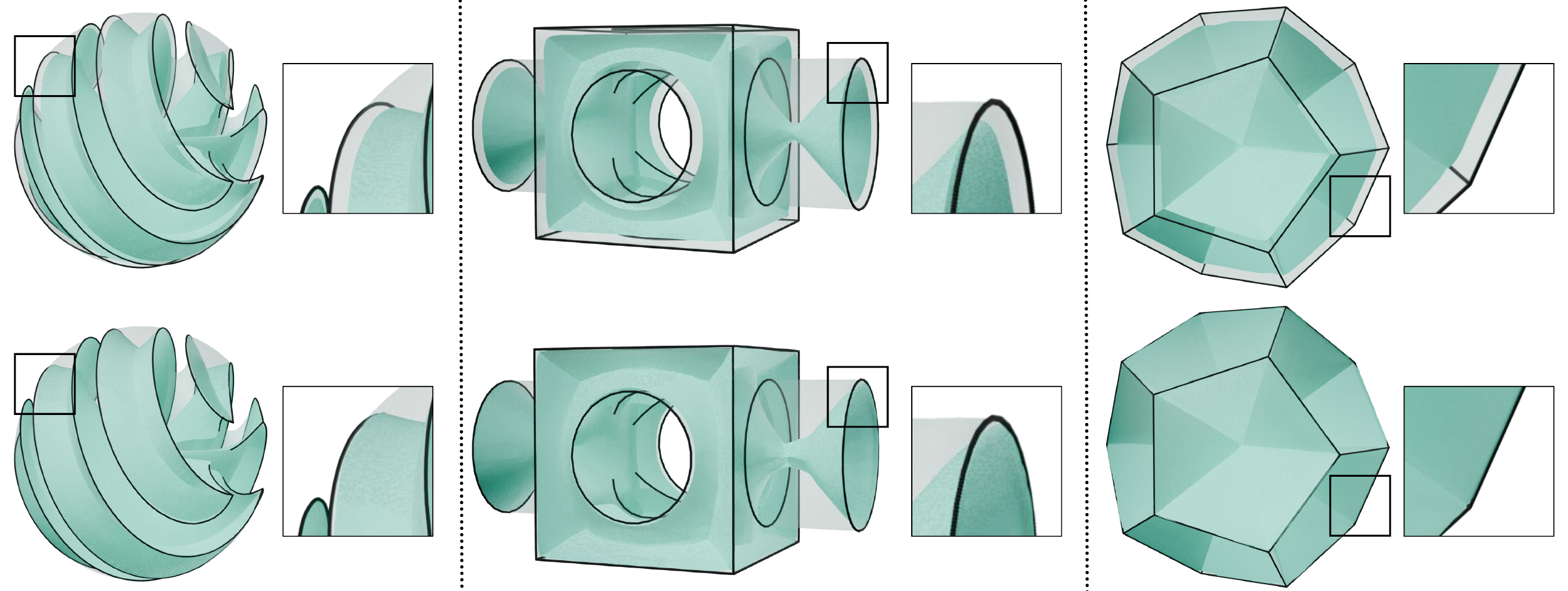}
    \caption{Sharp feature enhancement pulls the medial mesh, originally distant from sharp features, closer to them, significantly improving the quality of the extracted medial surfaces.}
    \label{fig:enhance_feature}
\end{figure}

\begin{figure*}[htb]
    \centering
    \includegraphics[width=0.98\linewidth]{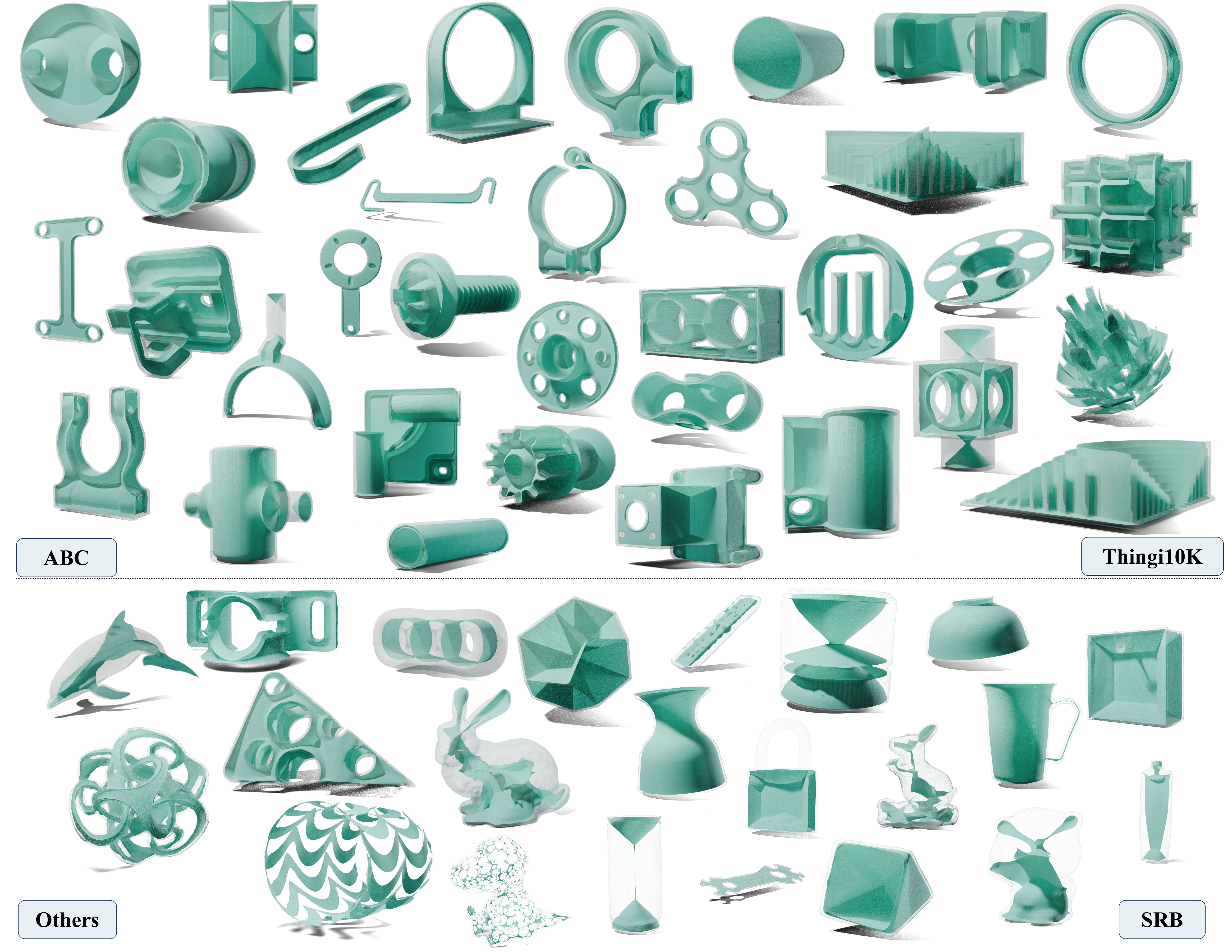}
    \caption{Gallery of results. A variety of challenging and classical models are shown, including examples from commonly used datasets such as Thingi10k~\cite{zhou2016thingi10k}, ABC~\cite{koch2019abc}, and SRB~\cite{huang2024surface}. Unless otherwise stated, all results are produced using the default parameter setting for consistency, with the feature extension option disabled; see Figure~\ref{fig:enhance_feature} for results with feature extension enabled.}
    \label{fig:gallery}
\end{figure*}

\subsubsection{Feature Priors} \label{subsubsec:featurepriors}
Sharp feature consolidation is an optional refinement procedure designed to improve accuracy around sharp regions in CAD-like models. When enabled, this module refines the learned fields in the vicinity of sharp features. To assess its effectiveness, we first compute a coarse medial axis from the input point cloud and identify prominent feature points. These feature points are then incorporated as additional constraints during joint training of the fields. This sharp feature enhancement yields a more precise medial axis that extends to feature lines, as demonstrated in Figure~\ref{fig:enhance_feature}.

\begin{table}[htb]
\centering
\begin{tabular}{@{}l|ll@{}}
    \toprule
    Datasets & CD (10$^{-3}$) & HD (10$^{-3}$) \\ 
    \midrule
    ABC~\cite{koch2019abc}       & 1.6004 & 20.3596  \\ 
    Thingi10K~\cite{zhou2016thingi10k} & 2.6029 & 18.9215  \\ 
    SRB~\cite{huang2024surface}       & 2.1010 & 15.8574 \\ 
    \bottomrule
\end{tabular}
    \caption{For three datasets with diverse inputs, this table collects the average Chamfer distance (CD, 10$^{-3}$) and Hausdorff distance (HD, 10$^{-3}$) of surfaces reconstructed from our resulting medial axes.}
    \label{tab:dataset-accuracy}
\end{table}

\begin{table*}[]
\scalebox{0.9}{
\begin{tabular}{@{}c|ccccccc@{}}
\toprule
                                                                  & \multicolumn{4}{c}{Input}                                                                                                                                                                                                                           & \multirow{2}{*}{Output} & \multirow{2}{*}{Homotopy} & \multirow{2}{*}{Compactness} \\ \cmidrule(lr){2-5}
                                                                  & \begin{tabular}[c]{@{}c@{}}watertight\\ mesh\end{tabular} & \begin{tabular}[c]{@{}c@{}}mesh with\\ defects\end{tabular} & \begin{tabular}[c]{@{}c@{}}dense\\ point cloud\end{tabular} & \begin{tabular}[c]{@{}c@{}}sparse\\ point cloud\end{tabular} &                         &                           &                              \\ \midrule
PC~\cite{amenta2001power}                   & -                                                         & -                                                          & $\checkmark$                                                & $\times$                                                     & Medial mesh             & Conditional               & $\times$                     \\
SAT~\cite{miklos2010discrete}               & $\checkmark$                                              & -                                                          & $\checkmark$                                                & $\times$                                                     & Medial mesh             & Conditional               & $\times$                     \\
Q-MAT~\cite{li2015q}                        & $\checkmark$                                              & -                                                          & $\times$                                                    & $\times$                                                     & Skeletal mesh           & Conditional               & $\checkmark$                 \\
VC~\cite{yan2018voxel}                      & $\checkmark$                                              & $\times$                                                   & $\times$                                                    & $\times$                                                     & Medial mesh             & Conditional               & $\times$                     \\
CoverageAxis++~\cite{wang2024coverage}      & $\checkmark$                                              & -                                                          & $\checkmark$                                                & $\times$                                                     & Skeletal mesh           & No                        & $\checkmark$                 \\
MATFP~\cite{wang2022computing}              & $\checkmark$                                              & $\times$                                                   & $\times$                                                    & $\times$                                                     & Medial mesh             & No                        & $\checkmark$                 \\
MATTopo~\cite{wang2024mattopo}              & $\checkmark$                                              & $\times$                                                   & $\times$                                                    & $\times$                                                     & Medial mesh             & Yes                       & $\checkmark$                 \\
DPC~\cite{Dpoints15}                        & -                                                         & -                                                          & $\checkmark$                                                & $\checkmark$                                                 & Meso-skeleton           & -                         & -                            \\ \midrule
Point2Skeleton~\cite{lin2021point2skeleton} & -                                                         & -                                                          & $\checkmark$                                                & $\checkmark$                                                 & Skeletal mesh           & No                        & $\checkmark$                 \\
NeuralSkeleton~\cite{Clemot2023neural}      & -                                                         & -                                                          & $\checkmark$                                                & $\checkmark$                                                 & Skeletal mesh           & No                        & $\checkmark$                 \\
Ours                                                              & $\checkmark$                                              & $\checkmark$                                               & $\checkmark$                                                & $\checkmark$                                                 & Medial membrane         & Conditional               & $\checkmark$                 \\ \bottomrule
\end{tabular}
}
\caption{Summary of representative 3D medial axis methods.}
\label{tab:method-comparison}
\end{table*}

\begin{table*}[htb]
\scalebox{0.73}{
\begin{tabular}{@{}c|clcllcllclccllcllclcc@{}}
\toprule
               & \multicolumn{3}{c}{bunny}                                                      & \multicolumn{3}{c}{kitten}                                                   & \multicolumn{3}{c}{holes}                                                    & \multicolumn{3}{c}{cheeze}                                  & \multicolumn{3}{c}{trim-star}                                                & \multicolumn{3}{c}{block}                                                    & \multicolumn{3}{c}{sharp-sphere}        \\ \cmidrule(l){2-22} 
               & CD                       & \multicolumn{1}{c}{HD} & \multicolumn{1}{c|}{$\chi$}    & \multicolumn{1}{c}{CD} & \multicolumn{1}{c}{HD} & \multicolumn{1}{c|}{$\chi$}    & \multicolumn{1}{c}{CD} & \multicolumn{1}{c}{HD} & \multicolumn{1}{c|}{$\chi$}    & \multicolumn{1}{c}{CD} & HD    & \multicolumn{1}{c|}{$\chi$}    & \multicolumn{1}{c}{CD} & \multicolumn{1}{c}{HD} & \multicolumn{1}{c|}{$\chi$}    & \multicolumn{1}{c}{CD} & \multicolumn{1}{c}{HD} & \multicolumn{1}{c|}{$\chi$}    & \multicolumn{1}{c}{CD} & HD     & $\chi$    \\ \midrule
PC             & \multicolumn{1}{l}{0.54} & 25.76                  & \multicolumn{1}{c|}{225,049} & 0.83                   & 15.22                  & \multicolumn{1}{c|}{-12} & 0.38                   & 7.97                   & \multicolumn{1}{c|}{-6} & 4.56                   & 48.8  & \multicolumn{1}{c|}{-33} & 0.81                   & 17.48                  & \multicolumn{1}{c|}{-5} & 1.67                   & 32.84                  & \multicolumn{1}{c|}{-3} & 1.16                   & 40.01  & -3 \\
CoverageAxis++ & 10.26                    & 64.64                  & \multicolumn{1}{c|}{37}    & 24.64                  & 88.02                  & \multicolumn{1}{c|}{14}    & 36.02                  & 88.71                  & \multicolumn{1}{c|}{7}     & 28.17                  & 81.5  & \multicolumn{1}{c|}{-11}   & 26.46                  & 85.93                  & \multicolumn{1}{c|}{147}   & 27.71                  & 100.38                 & \multicolumn{1}{c|}{32}    & 30.38                  & 115.12 & 28    \\
Q-MAT          & 5.12                     & 127.83                 & \multicolumn{1}{c|}{1}     & 3.37                   & 15.75                  & \multicolumn{1}{c|}{0}     & 2.46                   & 9.36                   & \multicolumn{1}{c|}{-2}    & 2.05                   & 16.84 & \multicolumn{1}{c|}{35}    & 4.58                   & 34.14                  & \multicolumn{1}{c|}{0}     & 5.78                   & 62.16                  & \multicolumn{1}{c|}{3}     & 5.75                   & 67.37  & 2    \\
VC             & 16.37                    & 104.48                 & \multicolumn{1}{c|}{1}     & 2.77                   & 12.55                  & \multicolumn{1}{c|}{264}   & 2.21                   & 9.39                   & \multicolumn{1}{c|}{-2}    & \multicolumn{1}{c}{-}  & -     & \multicolumn{1}{c|}{1389} & 4.09                   & 26.71                  & \multicolumn{1}{c|}{3}     & 4.84                   & 25.22                  & \multicolumn{1}{c|}{-9}    & 7.06                   & 72.14  & 77    \\
SAT            & 12.46                    & 50.8                   & \multicolumn{1}{c|}{47}    & 12.93                  & 32.18                  & \multicolumn{1}{c|}{83}    & 7.34                   & 14.19                  & \multicolumn{1}{c|}{54}    & 3.37                   & 12.47 & \multicolumn{1}{c|}{38}    & 5.88                   & 14.26                  & \multicolumn{1}{c|}{72}    & 6.56                   & 18.95                  & \multicolumn{1}{c|}{118}   & 9.08                   & 45.89  & 63    \\
MATFP          & 1.94                     & 39.98                  & \multicolumn{1}{c|}{8}     & 0.79                   & 24.43                  & \multicolumn{1}{c|}{5}     & 0.49                   & 8.12                   & \multicolumn{1}{c|}{-2}    & 0.28                   & 10.22 & \multicolumn{1}{c|}{-19}   & 0.99                   & 15.88                  & \multicolumn{1}{c|}{0}     & 0.32                   & 9.63                   & \multicolumn{1}{c|}{-2}    & 0.68                   & 16.47  & 1     \\
MATTopo        & 2.01                     & 16.93                  & \multicolumn{1}{c|}{1}     & 1.61                   & 25.91                  & \multicolumn{1}{c|}{0}     & 0.8                    & 8.75                   & \multicolumn{1}{c|}{-2}    & 1.18                   & 10.71 & \multicolumn{1}{c|}{-19}   & 0.78                   & 8.71                   & \multicolumn{1}{c|}{0}     & 0.97                   & 24.1                   & \multicolumn{1}{c|}{-2}    & 1.7                    & 23.39  & 1     \\ \midrule
NeuralSkeleton & -                        & \multicolumn{1}{c}{-}  & \multicolumn{1}{c|}{1}    & \multicolumn{1}{c}{-}  & \multicolumn{1}{c}{-}  & \multicolumn{1}{c|}{2}    & \multicolumn{1}{c}{-}  & \multicolumn{1}{c}{-}  & \multicolumn{1}{c|}{-2}    & \multicolumn{1}{c}{-}  & -     & \multicolumn{1}{c|}{-11}   & \multicolumn{1}{c}{-}  & \multicolumn{1}{c}{-}  & \multicolumn{1}{c|}{0}   & \multicolumn{1}{c}{-}  & \multicolumn{1}{c}{-}  & \multicolumn{1}{c|}{-2}   & \multicolumn{1}{c}{-}  & -      & 1  \\ \midrule
Ours           & 1.86                   & 12.56                  & \multicolumn{1}{c|}{1}     & 1.58                   & 19.96                  & \multicolumn{1}{c|} {0}                          & 0.89                   & 9.74                  & \multicolumn{1}{c|} {-2}                         & 3.12                   & 32.58 & \multicolumn{1}{c|}{-19}                         & 0.86                   & 16.54                  & \multicolumn{1}{c|} 0                          & 0.95                   & 21.41                  & \multicolumn{1}{c|} {-2}                          & 1.67                   & 19.69  & 1      \\ \bottomrule
\end{tabular}}
\caption{Comparison of reconstructed surface accuracy and topological correctness among eight representative methods and ours. CD and HD are scaled by $10^{3}$. The Euler characteristic ($\chi$) reflects the topology of the computed medial structure. Both MATTopo and our method yield topologically correct results.}
\label{tab:compare-accuracy1}
\end{table*}

\begin{figure*}
    \centering
    \begin{overpic}[scale=0.125,unit=1mm]{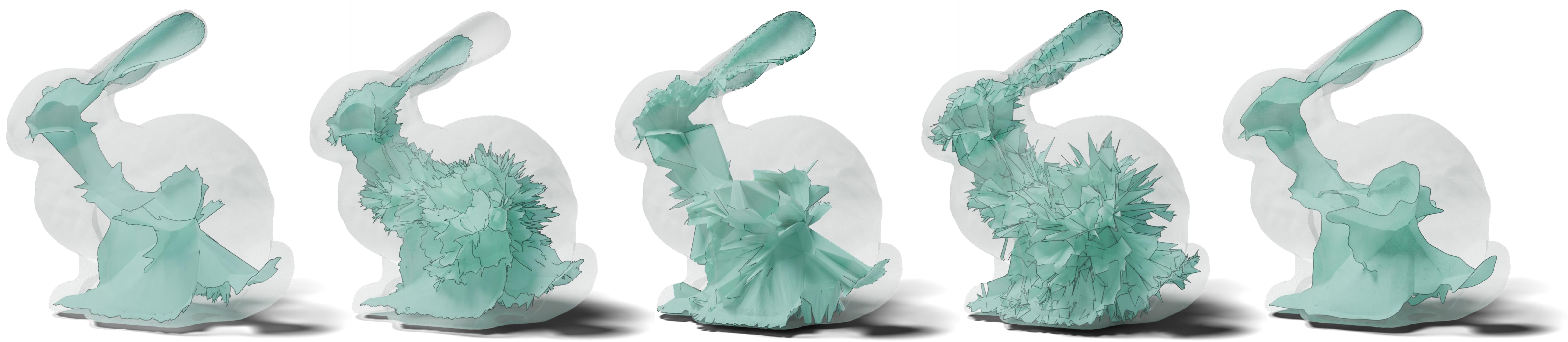}
     \put(8,-1){SAT}
    \put(28,-1){VC}
    \put(46,-1){MATFP}
    \put(65,-1){MATTopo}
    \put(87,-1){Ours}
    \end{overpic}
    \caption{Medial axes of the Bunny model computed by SAT~\cite{miklos2010discrete}, VC~\cite{yan2018voxel}, MATFP~\cite{wang2022computing}, MATTopo~\cite{wang2024mattopo}, and our method. The boundaries are visualized in black. Our method produces a more compact and smoother medial axis, whereas the other methods tend to exhibit excessive branching or jagged boundaries.    
    }
    \label{fig:bunny}
\end{figure*}

\subsubsection{Robustness}
Our method is robust and does not require a watertight manifold mesh or a densely sampled point cloud as input -- conditions often assumed by existing methods. In practice, the inputs can be a mesh with boundaries, a triangle soup, or even a sparse point cloud. We evaluate robustness on three benchmark datasets: ABC~\cite{koch2019abc}, Thingi10k~\cite{zhou2016thingi10k}, and Surface Reconstruction Benchmark (SRB)~\cite{huang2024surface}.
The ABC dataset contains numerous CAD-style meshes, while Thingi10K includes more complex geometric models that often have defects such as holes, self-intersections, and non-manifold structures. The SRB dataset consists of diverse point cloud inputs. For each dataset, we select the first 30-50 models for evaluation, and the statistical results are summarized in Table~\ref{tab:dataset-accuracy}. Additional qualitative examples using challenging point cloud inputs and commonly used 3D models are shown in Figure~\ref{fig:gallery}.

\subsection{Comparison}
We compare our method with ten representative approaches. Beyond geometric and topological accuracy, we also assess each method's ability to handle challenging cases such as thin sheets, sparse point clouds, and noisy data. A quantitative summary of the results is provided in Table~\ref{tab:method-comparison}.

\subsubsection{Geometry and Topology Accuracy}
We first evaluate the reconstruction quality and topological accuracy of the generated medial axes using toy examples, where ground-truths are available.
\begin{figure*}[htb]
    \centering
    \includegraphics[width=\linewidth]{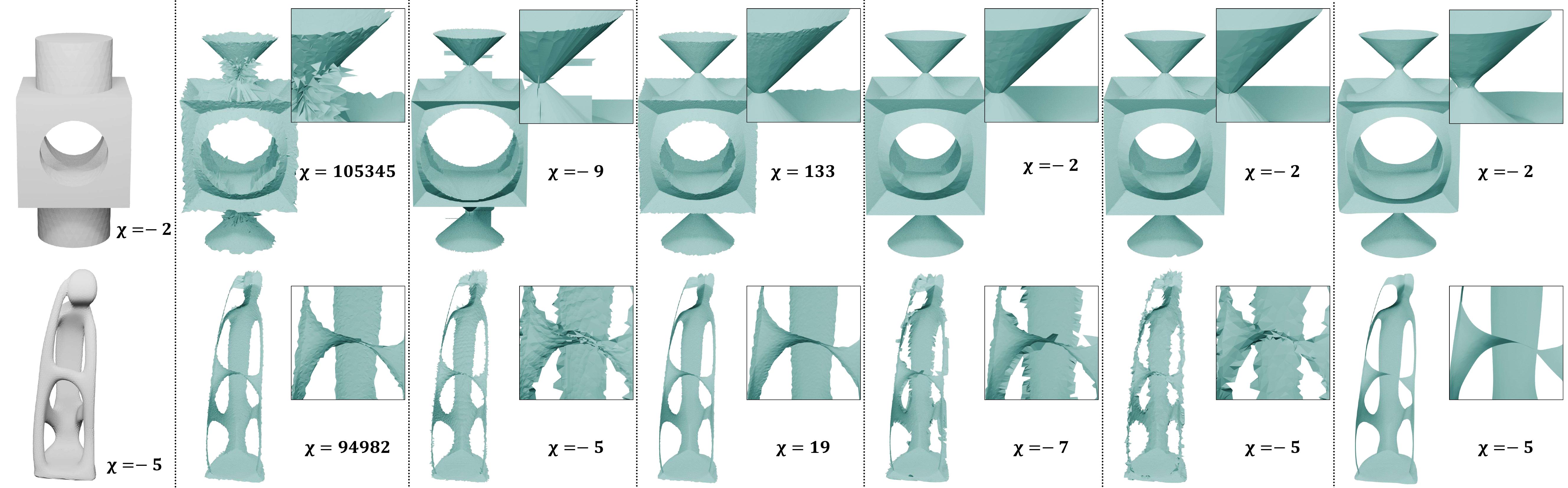}
    \makebox[0.135\linewidth][c]{Input}
     \makebox[0.135\linewidth][c]{PC}
       \makebox[0.135\linewidth][c]{VC}
        \makebox[0.135\linewidth][c]{SAT}
         \makebox[0.135\linewidth][c]{MATFP}
          \makebox[0.135\linewidth][c]{MATTopo}
           \makebox[0.135\linewidth][c]{Ours}
   \caption{Medial axes computed for both CAD and organic models, together with their corresponding Euler characteristics ($\chi$). Our method produces smooth and clean medial axes whose topology is consistent with that of the input shapes.}
    \label{fig:compare-topology}
\end{figure*}

The reconstruction accuracy
is measured using Chamfer distance (CD) and Hausdorff distance (HD) between the reconstructed surface mesh and the ground-truth mesh. It should be noted that the DPC method produces a meso-skeleton (i.e., a set of medial points combined with consolidated surface points rather than a connected medial mesh). Therefore, for fairness, it is excluded from the quantitative comparison. For all other methods, parameters were carefully tuned to obtain reasonable medial axis results.
The statistics reported in Table~\ref{tab:compare-accuracy1} show that, for organic models (the first three columns), the medial meshes computed by PC without pruning preserve finer structural details, leading to more accurate surface reconstruction. 
For CAD-like models (last three columns), feature-preserving methods such as MATFP and MATTopo achieve the best performance.
Although our method produces a relatively compact medial axis, it maintains stable and competitive geometric accuracy across both model types.

\re{In terms of topology, we conduct comparisons on both CAD and organic models (Figure~\ref{fig:compare-topology}), as well as on challenging models with thin structures  (Figure~\ref{fig:thin-sheet}).
Quantitative statistics are reported in Table~\ref{tab:compare-accuracy1}.

Both PC and SAT tend to suffer from the ``flat-tetrahedron problem'' when connecting medial points, often producing topologies that deviate from the input model. While VC can theoretically preserve topology, in practice, limited voxel resolution causes the voxelized shape to lose homotopy equivalence with the original geometry.
As illustrated in Figure~\ref{fig:thin-sheet}, even at resolution of $512^3$ or $1024^3$, VC often fails to capture thin structures, resulting in missing regions.
MATTopo, as an improved version of MATFP, effectively preserves topological correctness and performs well on CAD models and thin structures, though it may introduce redundant branches when applied to organic models.

}

\subsubsection{Smoothness}
This comparison is limited to methods capable of generating a ``clean'' medial mesh, defined as a non-manifold surface without internal cavities.  
The boundaries of these medial meshes are extracted by identifying triangles adjacent to only one face. \modi{For our method, the boundary of the medial membrane is determined using the  dihedral-angle criterion discussed in } Section~\ref{subsec:feature}. For CAD-like models, the medial meshes produced by MATFP and MATTopo ideally extend to sharp features, enabling superior smoothness controlled by the shape's sharp edges. However, for organic models, our method yields more stable and smoother boundaries, showing a clear advantage over the other approaches. A direct visual comparison of an organic model is presented in Figure~\ref{fig:bunny}, and detailed quantitative statistics are reported in Table~\ref{tab:smoothness}.

\begin{table}[htb]
\begin{tabular}{@{}l|lllll@{}}
\toprule
             & SAT  & VC   & MATFP   & MATTopo & Ours    \\ \midrule
Bunny        & 0.40 & 0.73 & 1.02    & 0.98    & 0.07    \\
Dodecahedron & 0.48 & 0.84 & 1.18e-6 & 1.11e-2 & 1.55e-2 \\ \bottomrule
\end{tabular}
\caption{Quantitative comparison of boundary smoothness for organic and CAD models, evaluated using our method, SAT~\cite{miklos2010discrete}, VC~\cite{yan2018voxel}, MATFP~\cite{wang2022computing}, and MATTopo~\cite{wang2024mattopo}.}
\label{tab:smoothness}
\end{table}

\subsubsection{Thin Structures}
\begin{figure}[htpb]
    \centering
 \includegraphics[width=0.95\linewidth]{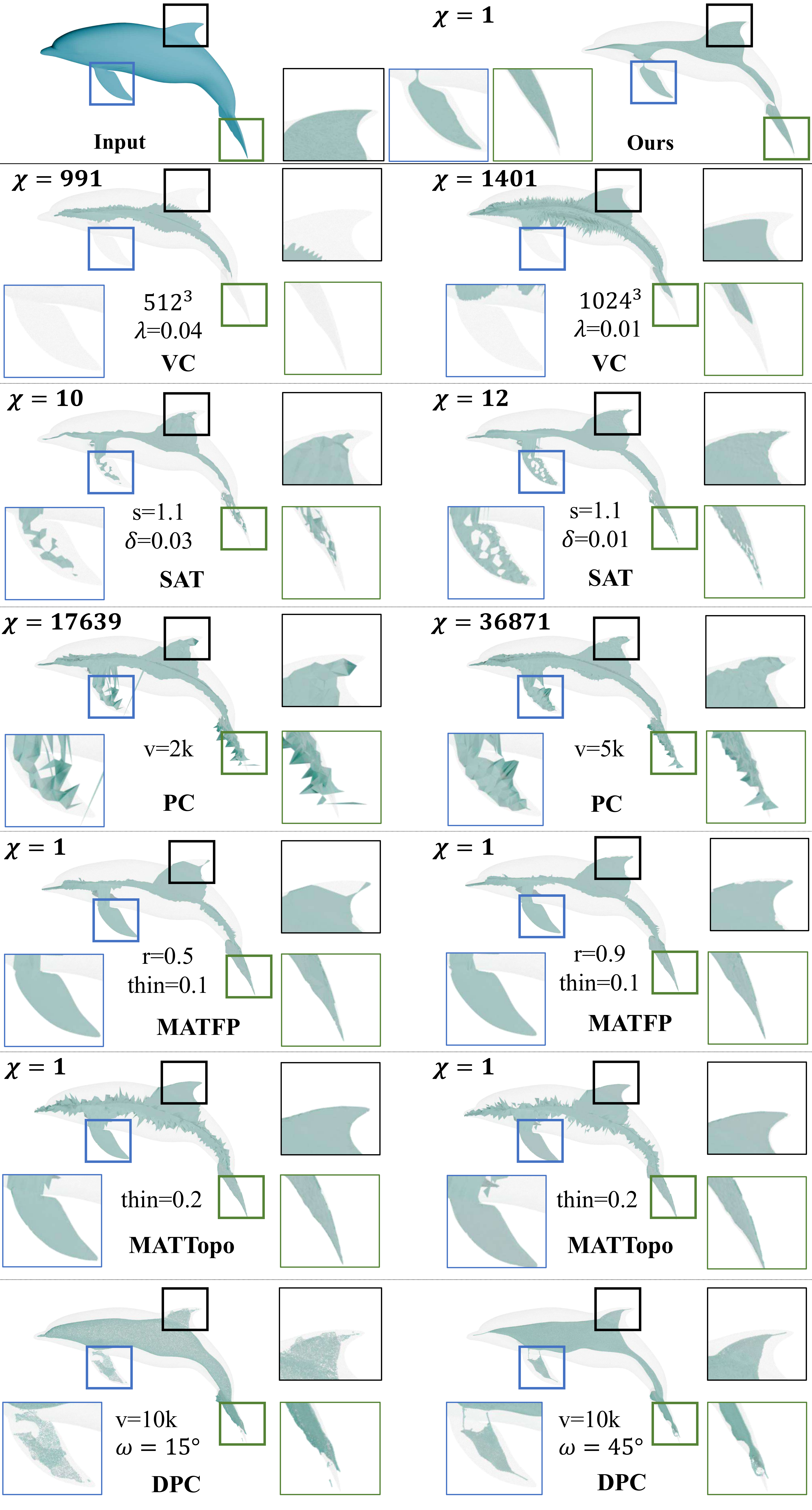}
    \caption{Qualitative comparison of our method with VC~\cite{yan2018voxel}, SAT~\cite{miklos2010discrete}, PC~\cite{amenta2001power}, MATFP~\cite{wang2022computing}, MATTopo~\cite{wang2024mattopo}, and DPC~\cite{Dpoints15} on the Dolphin model, which contains multiple thin structures on its tail and fins. For each baseline method, results are shown under two carefully tuned parameter settings chosen to achieve its best performance, accompanied by close-up views highlighting local details. Our method effectively handles these thin structures, producing a compact medial mesh that is superior in both geometry and topology.
    }
    \label{fig:thin-sheet}
\end{figure}

For models containing thin plates or tubes, most methods that rely on surface sampling or voxelization are \re{sensitive} to parameter variations.
As shown in Figure~\ref{fig:thin-sheet}, the input Dolphin model contains thin regions around its fins and flippers.
When the voxel resolution is insufficient, VC struggles to extract complete medial axes in these areas. 
Surface-sampling-based methods such as SAT, PC and MATFP also face challenges in achieving sufficiently dense sampling, leading to inaccuracies in both topology and geometry.
\modi{The meso-skeleton produced by DPC tends to shift away from the true center, and fewer medial points are generated in regions where the medial radius changes abruptly.} MATTopo, by progressively adding points and verifying topological correctness, can effectively explore thin-plate regions. It generates medial axes that are both geometrically and topologically accurate, and its results are relatively insensitive to parameter tuning. With default parameters, our method produces a well-behaved Q-MDF in thin-sheet regions, from which a compact medial axis can be easily extracted. The resulting medial axis is homotopy-consistent with the input shape and remains robust to variations in local thickness.

\begin{figure}[htp]
    \centering    \includegraphics[width=0.90\linewidth]{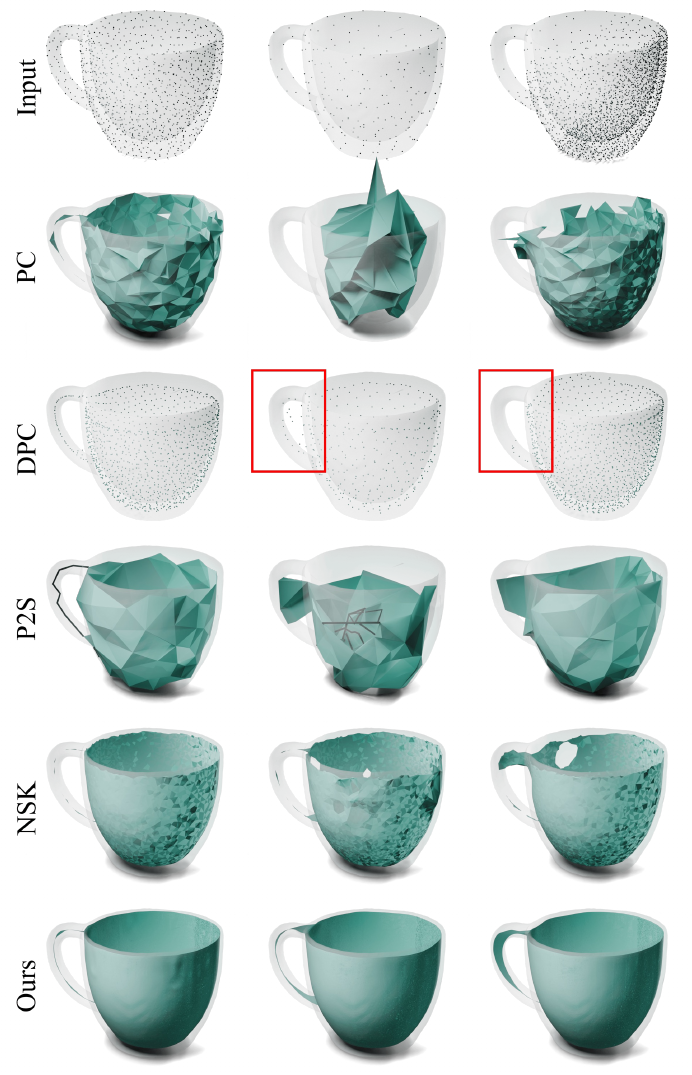}
    \caption{Comparison under varying point cloud densities. \modi{The first two columns show results from uniform sampling with different densities, while the last column illustrates the case with non-uniform density.}}
    \label{fig:sparse}
\end{figure}

\subsubsection{Sparse Inputs}
For most traditional methods that support point clouds as input, surface sampling density is a critical factor that strongly influences the quality of the resulting medial axes. Figure~\ref{fig:sparse} illustrates this effect using the Teacup model. The first two inputs are uniformly sampled point clouds with different densities, and the third is a point cloud with varying density. For PC, a representative  computational geometry method, as the input becomes sparser, both the centering and smoothness of the medial axis degrade. In extreme cases, identifying the interior poles becomes difficult.
DPC performs better than PC on sparse inputs, but the connection between the handle and the body of the teacup often exhibits geometric and topological inconsistencies. Similarly, when the input is sparse, Point2Skeleton struggles to identify interior points and form correct connections. NeuralSkeleton (NSK) benefits from learned shape inference and produces more reasonable medial meshes, though holes can appear when the input is extremely sparse. In contrast, our method demonstrates stable performance across all sparse-input cases. As shown in the three examples, the medial axes computed by our approach consistently preserve the topology of the original shape while maintaining good centering and smoothness.

\begin{figure}[htb]
    \centering
    \includegraphics[width=0.95\linewidth]{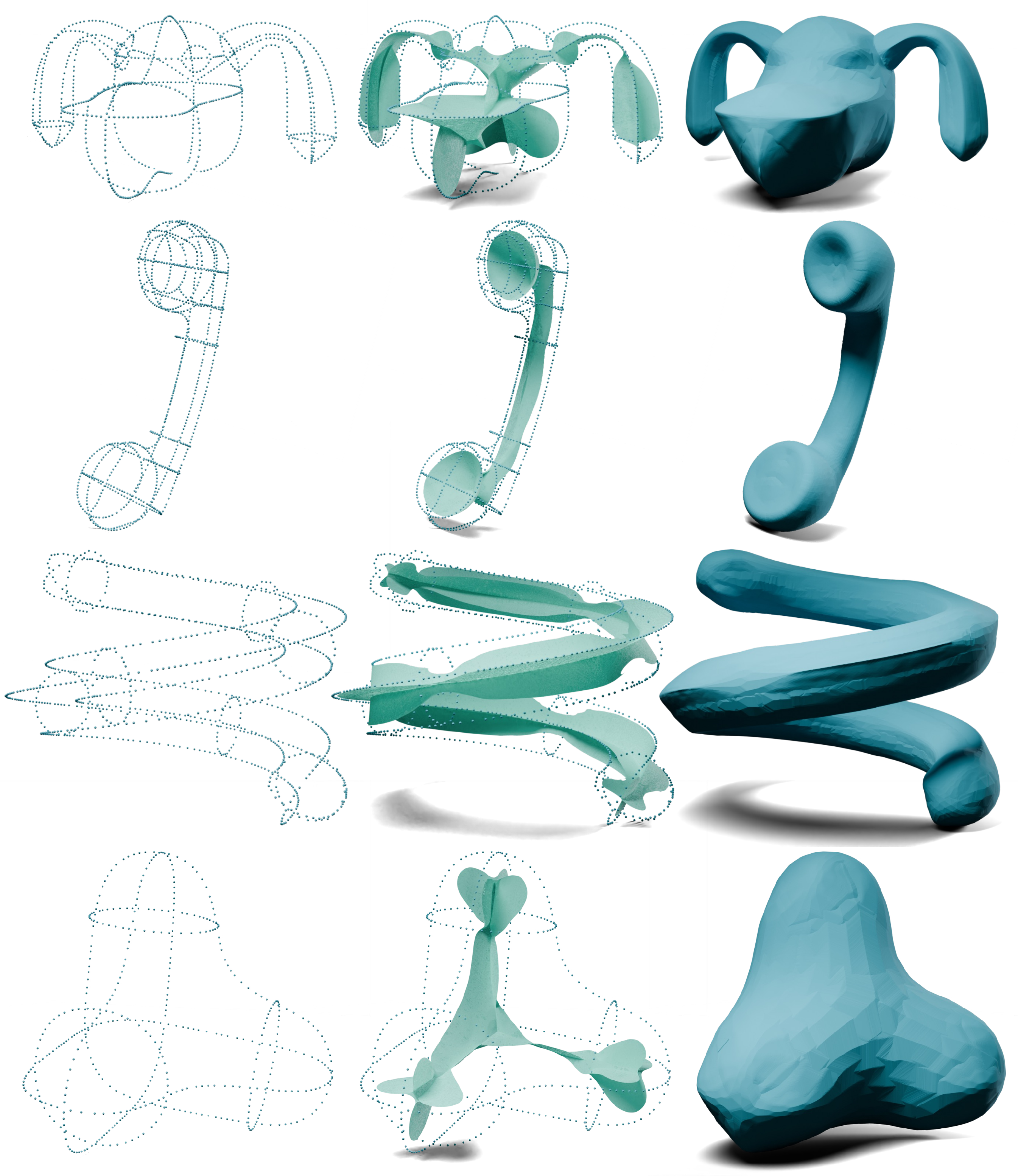}
    \caption{From left to right: wireframe inputs~\cite{huang2019variational}, computed medial axes, and  reconstructed surfaces using our method.}
    \label{fig:wireframe}
\end{figure}

In addition to uniformly sampled sparse inputs, our method can handle even more extreme scenarios, such as wireframe inputs. Figure~\ref{fig:wireframe} shows the extracted medial axis and reconstructed surface for the wireframe model from \cite{huang2019variational}.

\subsubsection{Incomplete Inputs}
\begin{figure*}[htb]
    \centering
    \includegraphics[width=0.95\linewidth]{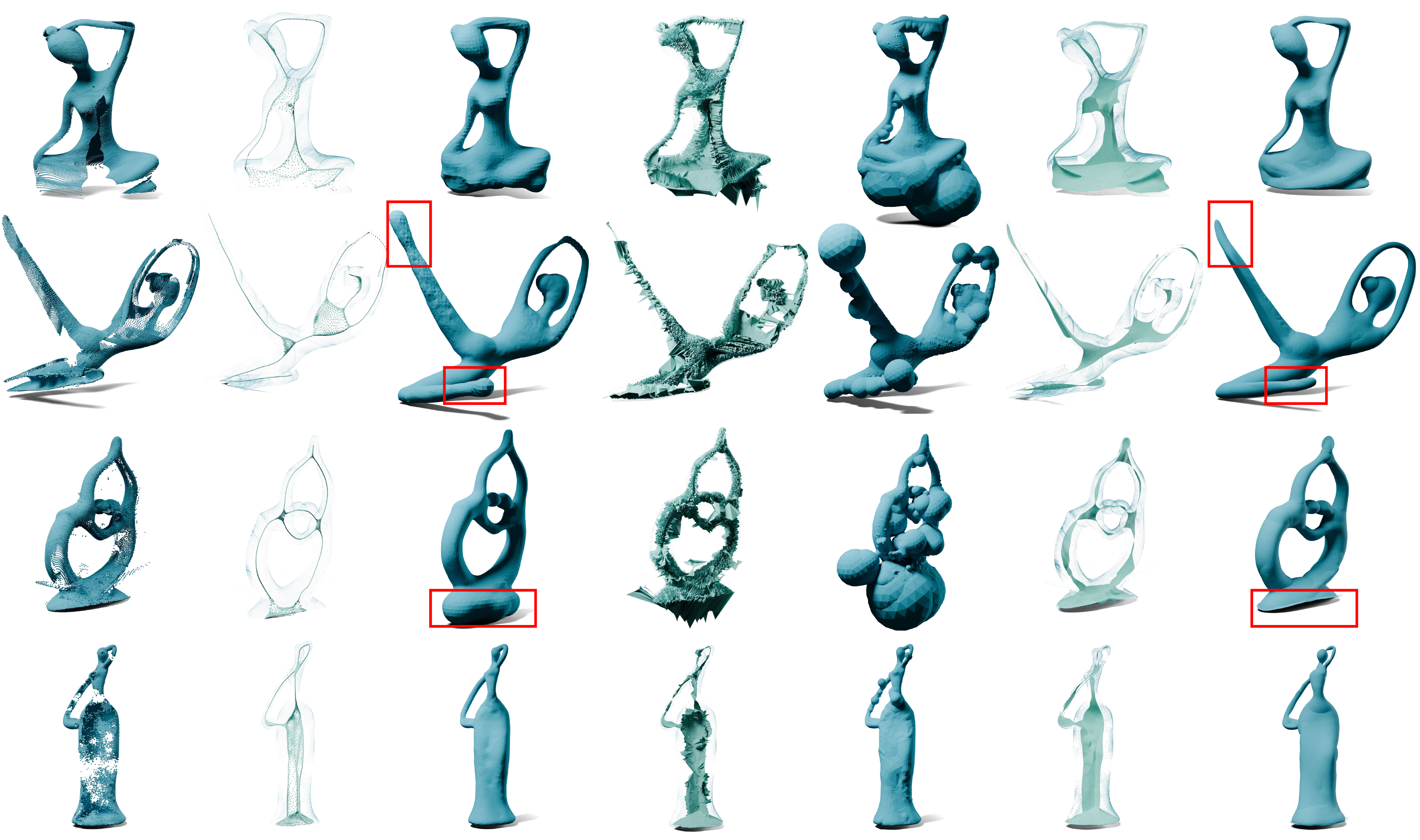}
     \makebox[0.13\linewidth][c]{Input}
      \makebox[0.13\linewidth][c]{Meso-skeleton}
    \makebox[0.13\linewidth][c]{DPC + Poisson}
    \makebox[0.13\linewidth][c]{PC MA}
    \makebox[0.13\linewidth][c]{PC Recons}
    \makebox[0.13\linewidth][c]{Our MA}
    \makebox[0.13\linewidth][c]{Our Recons}   
    \\
    \caption{Comparison of our method with DPC~\cite{Dpoints15} and PC~\cite{amenta2001power} on point clouds with partially missing data. From left to right: input point clouds; meso-skeletons and corresponding Poisson reconstructions generated from DPC surface samplings; medial meshes computed by PC and their reconstructed surfaces; and medial meshes produced by our method with their corresponding reconstructed surfaces. Despite the missing data, the input point clouds exhibit cyclide-like local geometry. Our method, designed to simplify and regularize the medial axis,  successfully reconstructs the medial structure from these incomplete inputs, yielding continuous and smooth surfaces that effectively fill in the missing regions.}
    \label{fig:huanghui_data}
\end{figure*}

When the input point cloud contains missing regions, obtaining a well-centered medial axis becomes challenging for most traditional methods. As shown in Figure~\ref{fig:huanghui_data}, when PC is applied to a point cloud with missing patches, the resulting medial axis bends toward the missing area or protrudes beyond the latent surface.
The corresponding radius values at these vertices are abnormally large, leading to poor surface reconstruction from the medial axis transform. DPC infers a complete surface sampling using localized symmetry priors, producing a centered point set known as the meso-skeleton.
It reconstructs the surface directly from the enhanced surface sampling via Poisson surface reconstruction~\cite{Kazhdan2013}. While the results appear plausible, when the missing data is extensive, DPC often introduces spherical artifacts. By contrast, our method leverages a neural implicit representation with localized symmetry patterns, producing a well-centered medial structure and a visually consistent, tightly reconstructed surface.

\subsubsection{Noise}
\begin{figure}[htb]
    \centering
    \includegraphics[width=0.95\linewidth]{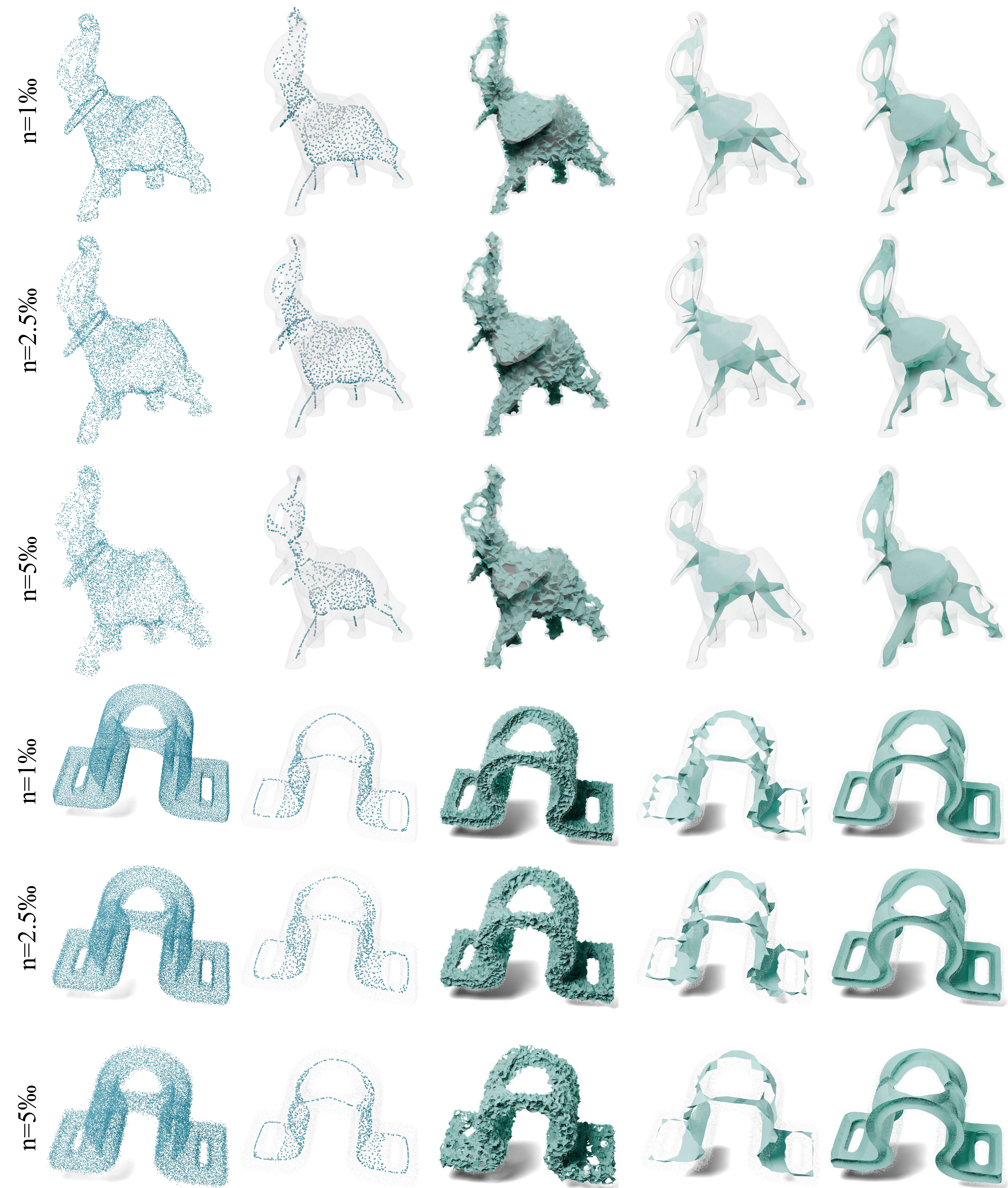}
    \\
   
       \makebox[0.19\linewidth][c]{Input}
       \makebox[0.19\linewidth][c]{DPC}
       \makebox[0.19\linewidth][c]{PC}
       \makebox[0.19\linewidth][c]{NSK}
    \makebox[0.19\linewidth][c]{Ours}
    \caption{Comparison of medial axis computation under different noise levels. For each model, the noise level increases from top to bottom. PC~\cite{amenta2001power} is highly sensitive to noise, whereas  DPC~\cite{Dpoints15}, NeuralSkeleton (NSK)~\cite{Clemot2023neural}, and our method exhibit improved robustness against noise.}
    \label{fig:noise}
\end{figure}

We also evaluate the robustness under varying noise levels. The test inputs consist of three point clouds sampled from the Elephant model with 0.1\%, 0.25\%, and 0.5\% Gaussian noise. As shown in Figure~\ref{fig:noise}, as the noise level increases, the medial meshes produced by PC increasingly protrude outside the surface, leading to topological errors such as holes or redundant connections. 
DPC, when provided with noisy point clouds and correct normals, produces visually reasonable shapes but suffers from off-center meso-skeletons.  
Both NeuralSkeleton and our method show strong resistance to noise, consistently generating compact and well-centered medial axes. For CAD-like models, however, NeuralSkeleton tends to oversimplify the medial structure, failing to capture fine geometric details.

\subsubsection{Real Scans}
\begin{figure}[htb]
    \centering
     \includegraphics[width=0.95\linewidth]{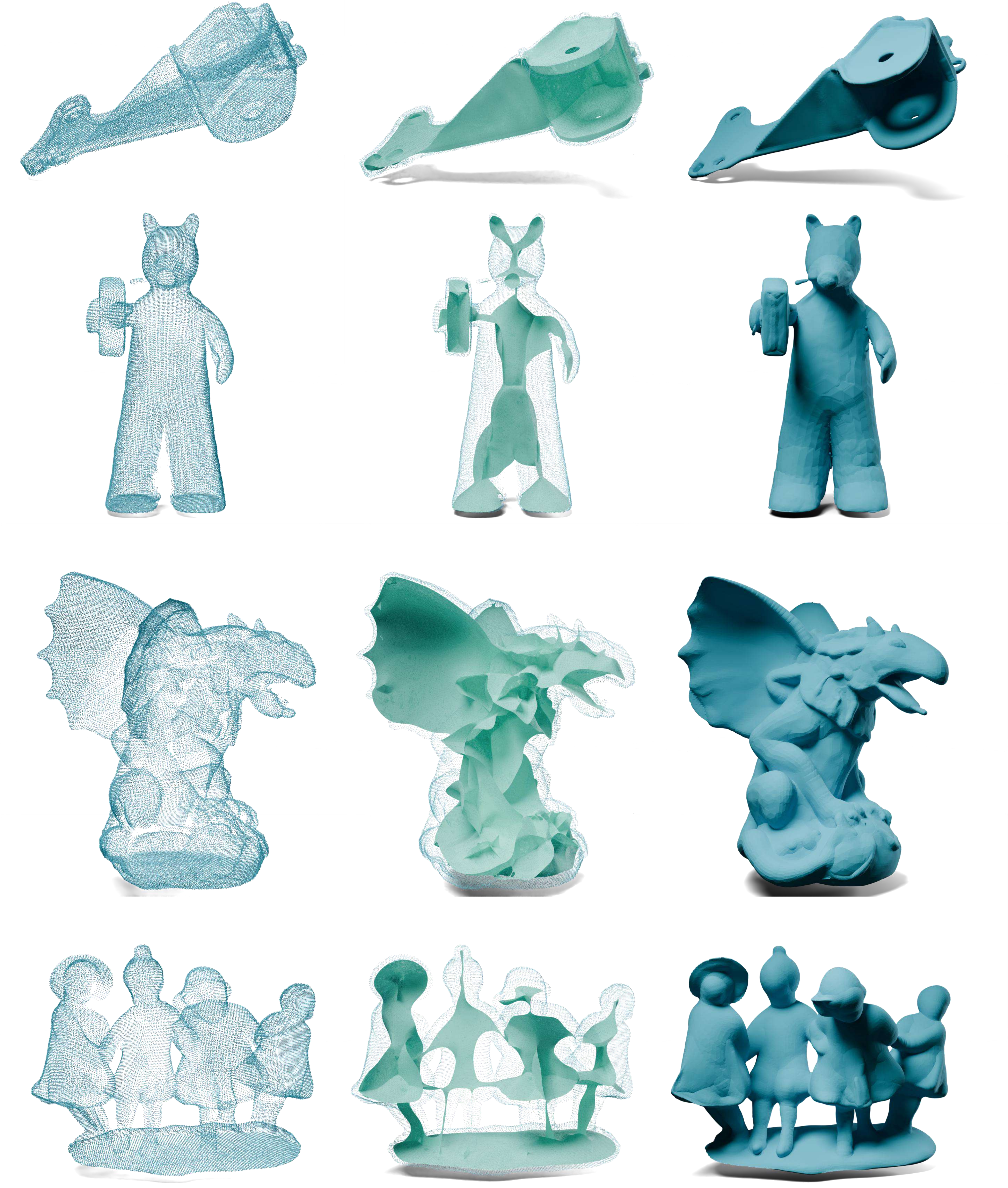}
     \makebox[0.29\linewidth][c]{Input}
    \makebox[0.29\linewidth][c]{Medial axes}
    \makebox[0.29\linewidth][c]{Reconstruction}\\
    \caption{Medial meshes and corresponding reconstruction results obtained by our method on real-world scan data.}
    \label{fig:real-scan}
\end{figure}

We further evaluate our method on real scan data from SRB~\cite{berger2013benchmark}.
These point clouds present challenges such as noise, misalignment, and missing patches. As shown in Figure~\ref{fig:real-scan}, the top row depicts a CAD model (Daratech) with a large area of missing points due to incomplete camera coverage.
Our method successfully handles this defect, producing a tight and coherent reconstruction. 
The remaining rows show organic models, for which our method generates compact medial axes with visually consistent topology and smooth reconstructed surfaces. 

\section{Conclusion and Future Work}
\label{sec:conclusion}
In this paper, we present a new method for computing the medial axis transform.  
Unlike traditional computational geometry approaches, which often have strict input requirements and suffer from numerical instability, our approach uses a neural implicit representation of the medial axis.  
This formulation enables for the robust extraction of a compact medial membrane mesh from various types of input data, including meshes, point clouds, and incomplete or noisy scans.

However, several issues remain to be addressed. First, the extracted membrane mesh is a double-layer structure with no explicit correspondence between the two overlapping patches, making the conversion to a single-layer non-manifold medial mesh challenging. Conventional double-layer peeling relies on a well-defined front/back separation of the surface~\cite{hou2023robust,dcudf2}; however, this separation becomes ill-defined in the
presence of non-manifold configurations, which are common in medial axis
meshes. As a result, a na\"ive slicing approach based solely on dihedral-angle detection (e.g., identifying fold-overs near $0$ or $2\pi$, depending on convention) is insufficient in these cases. For example, at a Y-shaped junction where three medial sheets meet, the extracted membrane contains seam edges with near-degenerate dihedral angles that can be detected and sliced. Yet, because there is no unique two-sided front/back separation at such junctions, slicing along these seams typically decomposes the membrane into three disconnected components rather than yielding two single-layered meshes. Thus, developing a principled, junction-aware strategy to convert the double-layer membrane into a single non-manifold medial mesh remains an interesting direction for future work~\cite{chen2025mind}. Second, our current implementation uses an MLP to train MF and SDF together, which is computationally expensive and limits practicality; replacing the backbone with multi-resolution hash/grid features and a small MLP head (Instant-NGP~\cite{mueller2022instant} style) could greatly improve performance. Finally, to enhance topological accuracy, our current implementation extracts the initial envelope using fixed-resolution Marching Cubes. For highly complex geometry, achieving reliable topology may require very high grid resolutions (e.g., $512^3$ or above), which significantly increases both memory usage and runtime. Incorporating a local feature size check (i.e., the Q-MDF value at surface points) and adopting an adaptive Marching Cubes scheme could reduce this issue by refining only where needed.

\section*{Acknowledgments}
We thank the anonymous reviewers for their constructive feedback, which has helped us significantly improve the quality of this paper.
Special thanks go to Reviewer~2 for providing highly detailed and insightful comments. This work was supported in part by the Ministry of Education, Singapore, under its Academic Research Fund Grant (RT19/22), the RIE2020 Industry Alignment Fund–Industry Collaboration Projects (IAF-ICP) Funding Initiative, cash and in-kind contribution from the industry partner(s), and the Research Projects of ISCAS (ISCAS-JCMS-202303, ISCAS-ZD-202401, ISCAS-JCZD-202402 \& ISCAS-JCMS-202403).

\bibliographystyle{ACM-Reference-Format}
\bibliography{sample-base}





\end{document}